\documentclass{article}

\PassOptionsToPackage{numbers, compress}{natbib}
\usepackage[preprint]{neurips_2024}

\usepackage[utf8]{inputenc} % allow utf-8 input
\usepackage[T1]{fontenc}    % use 8-bit T1 fonts
\usepackage{hyperref}       % hyperlinks
\usepackage{url}            % simple URL typesetting
\usepackage{booktabs}       % professional-quality tables
\usepackage{amsfonts}       % blackboard math symbols
\usepackage{nicefrac}       % compact symbols for 1/2, etc.
\usepackage{microtype}      % microtypography
\usepackage{xcolor}         % colors
\usepackage{graphicx}
\usepackage{amsmath}
\usepackage{listings}
\usepackage{xcolor}
\usepackage{enumitem}

\title{On the Opportunities of (Re)-Exploring Atmospheric Science by Foundation Models: A Case Study}

\author{%
Lujia Zhang\thanks{Equal contributions as first authors. Listing order is random.} \\
IIP, HKUST \\
\texttt{lzhangcq@connect.ust.hk}
\And 
Hanzhe Cui\footnotemark[1] \\
CE, HKUST\\
\texttt{hcuiah@connect.ust.hk}
\And 
Yurong Song\footnotemark[1] \\
CE, HKUST \\
\texttt{ysongbn@connect.ust.hk}
\And
Chenyue Li\footnotemark[1] \\
CSE, HKUST\\
\texttt{chenyue.li@connect.ust.hk}
\And
Binhang Yuan \\ 
CSE, HKUST\\
\texttt{biyuan@ust.hk}
\And
Mengqian Lu \\
CE, HKUST \\
\texttt{cemlu@ust.hk}
}

\begin{document}

\maketitle

\begin{abstract}
Most state-of-the-art AI applications in atmospheric science are based on classic deep learning approaches. However, such approaches cannot automatically integrate multiple complicated procedures to construct an intelligent agent, since each functionality is enabled by a separate model learned from independent climate datasets. The emergence of foundation models, especially multimodal foundation models, with their ability to process heterogeneous input data and execute complex tasks, offers a substantial opportunity to overcome this challenge. In this report, we want to explore a central question---\textit{how the state-of-the-art foundation model, i.e., GPT-4o, performs various atmospheric scientific tasks}. Toward this end, we conduct a case study by categorizing the tasks into four main classes, including climate data processing, physical diagnosis, forecast and prediction, and adaptation and mitigation. For each task, we comprehensively evaluate the GPT-4o's performance along with a concrete discussion. 
%We also list some future directions for the application of foundation models in atmospheric science.
We hope that this report may shed new light on future AI applications and research in atmospheric science. 

\end{abstract}

\section{Introduction}

%What is the problem?
The applications of artificial intelligence (AI) in atmospheric science have led to exciting and promising advances~\cite{sun2022review}. Different machine learning approaches have been applied in various tasks, including precipitation nowcasting~\cite{zhang2023skilful}, short- to medium-range weather prediction ~\cite{bi2023accurate,lam2023learning}, long-term climate pattern forecasting~\cite{ham2019deep,zhou2023self}, parameterization of atmospheric radiative transfer~\cite{yao2023physics}, and etc. 
Additionally, AI techniques have been instrumental in improving the accuracy of numerical modeling and enhancing the understanding of complex atmospheric processes~\cite{gregory2023deep,shen2023differentiable}. These advances illustrate the transformative potential of AI in atmospheric science, paving the way for more resilient and informed responses to climate challenges.

%Why is it hard? (E.g., why do naive approaches fail?)
%Why hasn't it been solved before? (Or, what's wrong with previous proposed solutions? How does mine differ?)
Most of the current applications of AI in atmospheric science rely on classic deep learning approaches, which typically consider separated datasets for specific problems, forcing the model to learn limited explicit or implicit statistical properties from these potentially homogeneous datasets~\cite{sun2022review}. 
However, practical deployment of climate-related workflow often requires careful integration of multiple complicated procedures, where an ideal AI agent should be capable of processing \textit{heterogeneous input data} and integrating a series of \textit{sophisticated reasoning procedures}.
Such requirements go beyond the scope of classic supervised or semi-supervised machine learning techniques --- addressing these challenges calls for the application of more advanced AI techniques that can dynamically adapt to diverse data types and execute complex, multi-step analytical tasks, thereby enhancing the robustness and applicability of AI in real-world atmospheric science scenarios.

%Consequently, developing a general model for atmospheric science, which can be widely applied to perform various tasks and execute step-by-step operations to address most questions, is highly valuable and warrants thorough exploration.

%Why is it interesting and important?

Recently, foundation models~\cite{bommasani2021opportunities} such as GPT-4~\cite{achiam2023gpt}, Gemini~\cite{team2023gemini}, Llama3~\cite{llama3}, and Mixtral~\cite{jiang2024mixtral} have essentially revolutionized the way state-of-the-art AI applications are developed and deployed.
The advancement of multimodal foundation models (FMs)~\cite{wu2023multimodal}, such as GPT-4o~\cite{GPT-4o}, Claude~\cite{anthropic2024claude} and Yi-VL~\cite{young2024yi} has further enhanced the ability of AI to process and understand information from diverse sources, such as text, time series, images, and videos. By integrating multiple representations of input data, these models achieve a more comprehensive understanding and generate more accurate and contextually relevant responses. This capability is particularly beneficial in fields requiring complex data interpretation, such as atmospheric science, where multi-modal FMs can seamlessly combine meteorological data, including satellite imagery, multiscale temperature or precipitation time series, and textual instructions to enable accurate weather predictions and complicated climate analysis.
Thus, we tend to expect that this great technique could create substantial new opportunities in atmospheric science based on the FM's capability to integrate complex multimodal heterogeneous inputs and perform advanced reasoning. 
Toward this end, we want to explore \textit{how the state-of-the-art foundation model, i.e., GPT-4o, performs various tasks within the field of atmospheric science}.

%GPT-4o (MNI) from OpenAI is recognized as one of the most powerful multi-modal large language models (MLLMs), capable of handling a wide range of inputs, including language, image, and spatial-temporal geoscience data. Through learning a large amount of statistical cases and academic papers, now it can help us handle data analysis and logical thinking to save the time and efforts of humans and provide new solutions to urgent challenges. By assimilating knowledge from extensive statistical cases and academic papers, it gains the ability to perform data analysis and logical reasoning, thereby not only employing established methods to save human time and effort but also offering innovative insights to address urgent challenges within the field of atmospheric science.

%What are the key components of my approach and results? Also include any specific limitations.
We answer this question by conducting a case study that explores the potential of state-of-the-art multi-modal FM to solve various climate tasks. Specifically, we categorize the tasks into four main classes to evaluate the performance of multi-modal FM comprehensively (Figure 1):

\begin{figure}[h]
    \centering
    \includegraphics[width=0.75\textwidth]{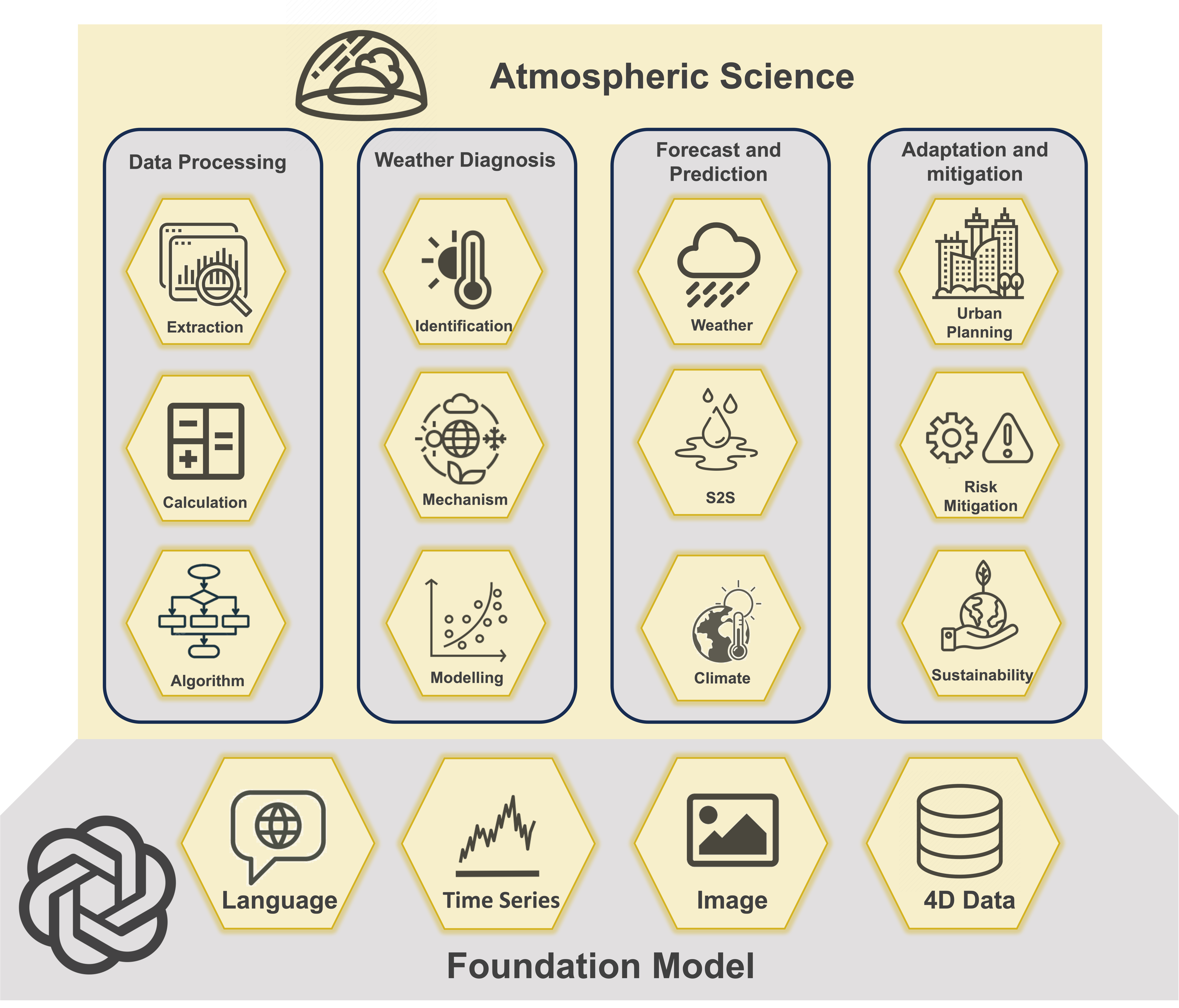} 
    \caption{Schematic of the relationship between the foundation model and different types of the tasks in the atmospheric science}
    \label{fig:1}
\end{figure}

\vspace{-0.5em}
\begin{itemize}[topsep=5pt, leftmargin=*]
    \item  \textbf{Climate data processing}. 
    Efficient processing of observational and modeling data is crucial for atmospheric scientific workflows. Tasks range from simple data extraction to advanced techniques like long-term trend calculation, linear regression coefficients, empirical orthogonal function (EOF) analysis, and K-means clustering, which require proficiency in various programming languages and consume significant research time. Delegating these tasks to the FM with strong mathematical reasoning and programming skills can greatly reduce scientists' workloads, allowing them to focus more on scientific inquiries.

    \item  \textbf{Physical diagnosis}. 
    Physical diagnosis in atmospheric science involves the analysis and interpretation of data to understand weather patterns, climate systems, and atmospheric processes, which leverage observational data, theoretical knowledge, and computational models to diagnose the atmosphere's state and behavior. An AI agent with a deep understanding of atmospheric dynamics and thermodynamics can greatly assist weather forecasters by handling extensive post-processing of weather maps, identifying potential outcomes of anomalous weather variables, and diagnosing extreme weather events. The ability of FM to handle basic statistical modeling, quantify relationships, predict future conditions, etc., can provide valuable support to atmospheric scientists in both research and operational tasks.
    
    \item  \textbf{Forecast and prediction}.
    %Predictions about possible future states of the weather or climate across various timescales have always been a central element of atmospheric science \cite{meehl2021initialized}. Short-term forecast alerts about impending weather hazards are crucial for preventing property loss and ensuring human safety. Forecasts on longer timescales, such as Subseasonal-to-Seasonal (S2S) and long-term, provide policymakers with valuable information regarding potential droughts or floods, enabling them to design appropriate strategies for agriculture and engineering \cite{cai2023anthropogenic}. Therefore, determining whether GPT-4o can assist in addressing this type of problem is incorporated in this report.
    Predictions of future weather and climate states on various time scales are central to atmospheric science~\cite{meehl2021initialized}. Short-term forecasts of imminent weather hazards are crucial to preventing property loss and ensuring human safety. Longer-term forecasts, such as subseasonal-to-seasonal and beyond, provide valuable information to policymakers to manage potential droughts or floods, helping to develop agricultural and engineering strategies~\cite{cai2023anthropogenic}. 

    \item \textbf{Adaptation and mitigation}.
    %In recent years, the frequency of natural disasters has noticeably increased, with various typhoons and extreme rainfall events causing significant damage. Concurrently, human activities, including carbon dioxide and air pollutant emissions, have exacerbated the greenhouse effect and deteriorated air quality, resulting in immeasurable harm. Consequently, it is crucial for governments to formulate effective policies to mitigate natural disasters, prevent economic and population losses, ensure timely post-disaster reconstruction and recovery, and reduce air pollution and greenhouse gas emissions. Addressing these issues is of paramount importance.
    Atmospheric science plays a crucial role in developing strategies to adapt and mitigate the impacts of climate change. Natural disasters such as typhoons and extreme rainfall have increased, causing significant damage. Human activities, such as carbon dioxide and air pollutant emissions, have worsened the greenhouse effect and air quality, causing further harm. Governments must formulate effective policies to mitigate natural disasters, prevent economic and population losses, ensure timely post-disaster recovery, and reduce air pollution and greenhouse gas emissions. 
    
\end{itemize}
\vspace{-0.5em}

In the rest of this report, we enumerate the case study output on each of the above tasks --- Section \ref{sec:data_process} on climate data processing, Section \ref{sec:phy_diag} on physical diagnosis, Section \ref{sec:pred} on forecast and prediction, and Section \ref{sec:adaptation} on climate adaptation and mitigation. Related works is shown in Section \ref{sec:relate}. At last, we conclude and discuss the results in Section \ref{sec:con}.

%\section{Categorization of Atmospheric Science Tasks}

\section{Climate Data Processing}
\label{sec:data_process}
The rapid development of atmospheric science has greatly benefited from the acquisition of massive amounts of data derived from various observation methods, like satellites, radar, ground stations, etc. This section aims to evaluate the capability of GPT-4o in assisting with three selected data processing tasks. Specifically, the accurate extraction of data, the correct execution of common calculations, and the application of typical algorithms are assessed in the following subsections.

\subsection{Information Extraction} \label{sec:2.1}
We begin by testing whether the latest version of GPT, GPT-4o, can read a commonly used file format in atmospheric science, netCDF4, and accurately extract the necessary information.

\begin{figure}[h!]
    \centering
    \includegraphics[width=0.7\textwidth]{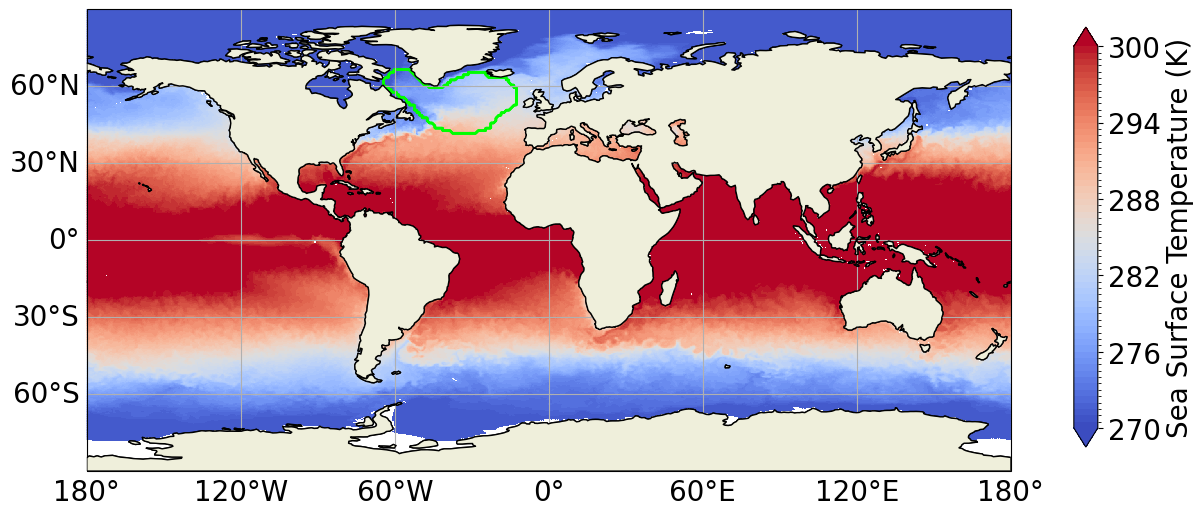} 
    \caption{Distribution of Global sea surface temperature Data on May 1, 2024, at 00:00:00. The green boundary indicates the region of the subpolar gyre. The data is sent to GPT-4o, and GPT-4o correctly processes this information. Refer to Section \ref{sec:2.1} for detailed discussions.
    }
    \label{fig:2}
\end{figure}

\noindent\textbf{Task inputs.}
We provide the sea surface temperature data for May 1, 2024, sourced from ERA5, along with the shapefile for the subpolar gyre (Figure \ref{fig:2}) along with the prompt:

\begin{quote}
\textcolor{cyan}{\texttt{"The first file is a netCDF4 file containing global sea surface temperature data. The variable name in this file is "sst," and its coordinates are longitude, latitude, and time. The second file is a netCDF4 file containing a mask. In this file, values equal to 1 represent the location of the Subpolar Gyre. Could you read these two files and tell me what the average sea surface temperature of the Subpolar Gyre is?"}}
\end{quote}

\noindent\textbf{GPT-4o output summary.} GPT-4o generates the synthesized code (Code Fragment \ref{code:2.1}) to conduct the analysis. 

\definecolor{codebg}{rgb}{0.95,0.95,0.95}
\lstset{
    backgroundcolor=\color{codebg},
    basicstyle=\ttfamily\small,
    columns=fullflexible,
    breaklines=true,
    captionpos=b,
    keepspaces=true,
    showspaces=false,
    showstringspaces=false,
    showtabs=false,
    tabsize=4,
    keywordstyle=\color{blue},
    commentstyle=\color{green!50!black},
    stringstyle=\color{red},
    frame=single,
    breakatwhitespace=false
}
\begin{lstlisting}[float=h!, language=Python, label=code:2.1, caption=GPT-4o synthesized code for extracting information from netCDF4 files.]
import xarray as xr
import numpy as np

# Open the netCDF files using xarray
sst_dataset = xr.open_dataset('/mnt/data/SST.nc')
sg_dataset = xr.open_dataset('/mnt/data/SG_region.nc')
sst = sst_dataset['sst']
sg_mask = sg_dataset['SG']

# Ensure the mask is broadcasted properly across all time steps if necessary
if sg_mask.ndim == 2:
    sg_mask = sg_mask.expand_dims(dim='time', axis=0)

# Apply the mask to the SST data
masked_sst = sst.where(sg_mask == 1)

# Calculate the average SST in the Subpolar Gyre region
average_sst_sg = masked_sst.mean().item()
\end{lstlisting}

The descriptive output is summarized as:

\begin{quote}
\textcolor{teal}{\texttt{The average sea surface temperature of the Subpolar Gyre is approximately 281.13 K.}}
\end{quote}

\noindent\textbf{Discussion.} We observe that the sea surface temperature is approximately 280 K (Figure \ref{fig:2}). Our manual calculations further show that the result from GPT-4o exactly matches the correct answer, demonstrating GPT-4o's capability to handle netCDF4 files and accurately extract data from the specified mask of the subpolar gyre. The success of this mission lays the foundation for subsequent analyses to perform more complex executions.

\subsection{Statistical Calculation}\label{sec:2.2}

It is an essential procedure for researchers to rely on various statistical calculations to obtain results. In this part, we specialize this type of task for GPT-4o to calculate the trend of sea surface temperature from 2001 to 2020.

\noindent\textbf{Task inputs.}
Similar to the previous task, we provide GPT-4o with netCDF4 data containing annual sea surface temperature values from 2001 to 2020 and instruct it to calculate the annual trend for each grid along with the prompt:

\begin{quote}

\textcolor{cyan}{\texttt{"The provided file contains annual sea surface temperature data from 2001 to 2020. Please calculate the trend of the sea surface temperature over this period and return the result in a netCDF4 file format."}}. 
\end{quote}

\begin{figure}[h]
    \centering
    \includegraphics[width=0.7\textwidth]{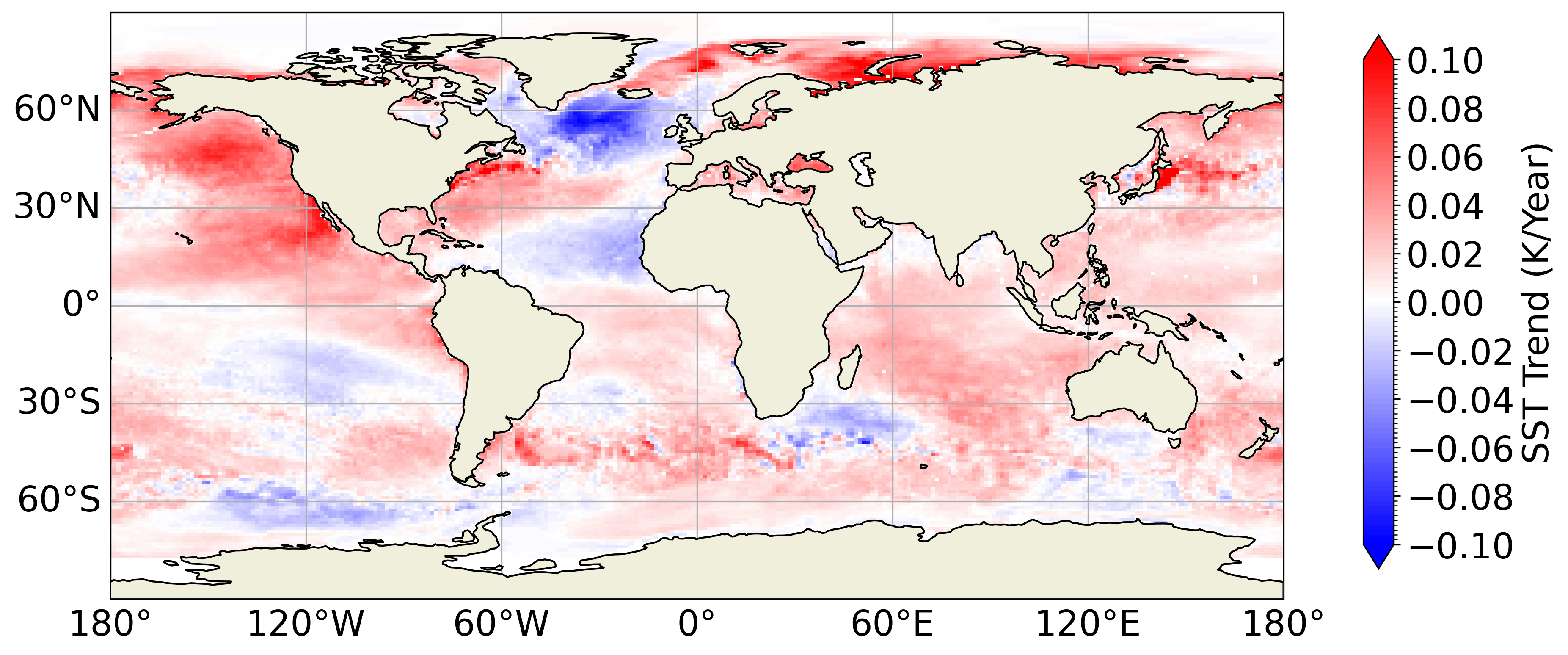} 
    \caption{Distribution of the sea surface temperature trend from 2001 to 2020. The plotted data is generated by GPT-4o, which correctly calculates the trend. Refer to Section \ref{sec:2.2}  for detailed discussions.}
    \label{fig:3}
\end{figure}

\noindent\textbf{GPT-4o output summary.} GPT-4o generates the synthesized code (Code Fragment \ref{code:2.2}) to conduct the analysis and generate the result.

\definecolor{codebg}{rgb}{0.95,0.95,0.95}
\lstset{
    backgroundcolor=\color{codebg},
    basicstyle=\ttfamily\small,
    columns=fullflexible,
    breaklines=true,
    captionpos=b,
    keepspaces=true,
    showspaces=false,
    showstringspaces=false,
    showtabs=false,
    tabsize=4,
    keywordstyle=\color{blue},
    commentstyle=\color{green!50!black},
    stringstyle=\color{red},
    frame=single,
    breakatwhitespace=false
}
\begin{lstlisting}[float=h!, language=Python, label=code:2.2, caption=GPT-4o synthesized code for extracting information from netCDF4 files.]
import xarray as xr
import numpy as np
import matplotlib.pyplot as plt
from scipy.stats import linregress

# Open the provided NetCDF file
file_path = '/mnt/data/SST_trend.nc'
dataset = xr.open_dataset(file_path)
years = dataset['year'].values
latitudes = dataset['lat'].values
longitudes = dataset['lon'].values
sst = dataset['SST'].values

# Calculate the trend over the years
trend = np.zeros((len(latitudes), len(longitudes)))

for i in range(len(latitudes)):
    for j in range(len(longitudes)):
        y = sst[:, i, j]
        if np.any(np.isnan(y)):
            trend[i, j] = np.nan
        else:
            slope, intercept, r_value, p_value, std_err = linregress(years, y)
            trend[i, j] = slope

# Create a new NetCDF file with the trend data
output_path = '/mnt/data/SST_trend_output.nc'
trend_dataset = xr.Dataset(
    {
        'SST_trend': (['lat', 'lon'], trend)
    },
    coords={
        'lat': latitudes,
        'lon': longitudes
    }
)
trend_dataset.to_netcdf(output_path)
\end{lstlisting}

The result is correctly generated and is manually shown in Figure \ref{fig:3}.

\noindent\textbf{Discussion.} GPT-4o correctly identifies the NaN values in the sea surface temperature data and returns the annual trend calculated from linear regression (Code Fragment \ref{code:2.2}). The detailed procedure used by GPT-4o to perform the calculation instills trust among scientists and lays the foundation for more advanced computational applications.

\subsection{Classical Algorithm}\label{sec:2.3}

We proceed to test whether GPT-4o could perform more advanced calculation, like the EOF analysis, which is a very classical and widely used algorithm in atmospheric science. 

\noindent\textbf{Task inputs.}
We utilize the monthly sea level pressure from 1980 to 2019 within the region 30°N-80°N, 80°E-130°W  for the EOF test along with the prompt:
 
\begin{quote}
\textcolor{cyan}{\texttt{"The provided file contains monthly sea level pressure data from 2001 to 2019.  Please help me do the EOF of this Sea level pressure file, the cos value of the latitude should be the weight of the EOF decomposition. Please return the first three EOF patterns the corresponding PCs and variance fraction to me through one netCDF4 file. Please remove the climatology first to do the EOF analysis."}}
\end{quote}

\begin{figure}[h]
    \centering
    \includegraphics[width=0.7\textwidth]{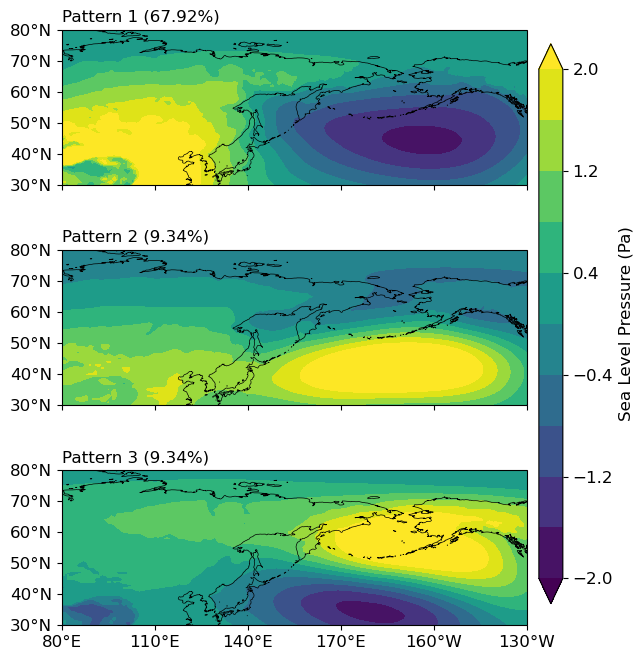} 
    \caption{
    Three EOF patterns generated by GPT-4o. Shading represents the sea level pressure for each pattern. The variance fraction of each EOF pattern is indicated at the top of each panel. The plotted data is generated by GPT-4o. Refer to Section \ref{sec:2.3}  for detailed discussions.
    }
    \label{fig:4}
\end{figure}

\noindent\textbf{GPT-4o output summary.} GPT-4o generates the synthesized code (Code Fragment \ref{code:2.3}) to conduct the analysis and generate the result. 

\definecolor{codebg}{rgb}{0.95,0.95,0.95}
\lstset{
    backgroundcolor=\color{codebg},
    basicstyle=\ttfamily\small,
    columns=fullflexible,
    breaklines=true,
    captionpos=b,
    keepspaces=true,
    showspaces=false,
    showstringspaces=false,
    showtabs=false,
    tabsize=4,
    keywordstyle=\color{blue},
    commentstyle=\color{green!50!black},
    stringstyle=\color{red},
    frame=single,
    breakatwhitespace=false
}
\begin{lstlisting}[language=Python, label=code:2.3, caption=GPT-4o Synthesized Code for EOF Analysis of sea level pressure.]
import xarray as xr
import numpy as np
from sklearn.decomposition import PCA
# Load the data
file_path = '/mnt/data/SLP_file.nc'
ds = xr.open_dataset(file_path)
# Adjust the process for applying weights properly
# Calculate the climatology
climatology = ds.mean('time')

# Remove the climatology to get anomalies
anomalies = ds - climatology

# Calculate the weights (cosine of latitude)
latitudes = ds['latitude']
weights = np.cos(np.deg2rad(latitudes))

# Expand the weights to match the dimensions of anomalies
weights_expanded = xr.DataArray(weights, dims=['latitude']).broadcast_like(anomalies)

# Apply the weights to anomalies
weighted_anomalies = anomalies * weights_expanded

# Perform PCA (EOF analysis)
anomalies_flat = weighted_anomalies.stack(points=('latitude', 'longitude'))
# Convert anomalies_flat to a DataArray before applying PCA
anomalies_flat_da = anomalies_flat.to_array().squeeze()

# Perform PCA (EOF analysis)
pca = PCA(n_components=3)
pcs = pca.fit_transform(anomalies_flat_da)

# Reshape EOFs to original spatial dimensions
eofs = pca.components_.reshape((3, ds.dims['latitude'], ds.dims['longitude']))

# Get variance fractions
variance_fractions = pca.explained_variance_ratio_

# Create xarray DataArrays for EOFs and PCs
eofs_da = xr.DataArray(eofs, dims=['mode', 'latitude', 'longitude'], name='EOFs')
pcs_da = xr.DataArray(pcs, dims=['time', 'mode'], name='PCs')
variance_da = xr.DataArray(variance_fractions, dims=['mode'], name='variance_fraction')

# Combine into a single dataset
final_ds = xr.Dataset({'EOFs': eofs_da, 'PCs': pcs_da, 'variance_fraction': variance_da})

# Save to netCDF4
output_file = '/mnt/data/EOF_SLP_analysis.nc'
final_ds.to_netcdf(output_file)
\end{lstlisting}

The result is correctly generated and is shown in Figure \ref{fig:4}.

\noindent\textbf{Discussion.} GPT-4o utilizes the Python library "sklearn" to execute the calculation (see Code Fragment \ref{code:2.3}), successfully obtaining the correct three EOF patterns and the corresponding variance fractions. This example further demonstrates GPT-4o's potential to solve more complex problems for atmospheric scientists by using widely acknowledged and accurate code to process the provided data. Overall, GPT-4o performs excellently on all three kinds of tasks in this section. This achievement is attributed to the latest version of GPT, which includes the capability to directly run code with netCDF4 data on its own servers—an ability not available in GPT-4 and GPT-3.5. Therefore, GPT-4o is poised to assist scientists in managing many tedious and repetitive tasks in daily research.

\section{Physical Diagnosis}
\label{sec:phy_diag}
Physical diagnosis is an important part of understanding the underlying phenomena and mechanisms behind atmospheric data and aids in further forecasting and prediction. This section aims to evaluate the capability of GPT-4o in assisting with three selected physical diagnosis tasks. Specifically, extreme weather detection, inference from meteorological variable fields, and statistical modeling are assessed in the following subsections. Traditional manual interpretation of weather maps for judgment is labor-intensive and time-consuming, and it is prone to omissions. Considering these challenges, we test GPT-4o's capability in diagnosing weather maps for an extreme rainfall event in Henan province.

\subsection{Extreme Weather Detection}\label{sec:3.1}
Identifying extreme weather events, such as heavy rainfall, tropical cyclones, and heatwaves, is crucial for conducting hazard assessments, risk analysis, and tracking the rapid development of hazards. In this test, we utilize the total precipitation netCDF4 file from July 20, 2021, for the East Asia region. 

\noindent\textbf{Task inputs.}
We provide the precipitation data for July 20, 2021, sourced from ERA5, along with the prompt:

\begin{quote}
\textcolor{cyan}{\texttt{"Attached file is a precipitation netCDF4 file (unit: m) for a certain day. Could you analyze this data and identify all the locations of the possible extreme precipitation (95 percentile) that occurred on that day and then highlight all the locations using a red box in the figure?"}}
\end{quote}

\begin{figure}
    \centering
    \includegraphics[width=0.8\textwidth]{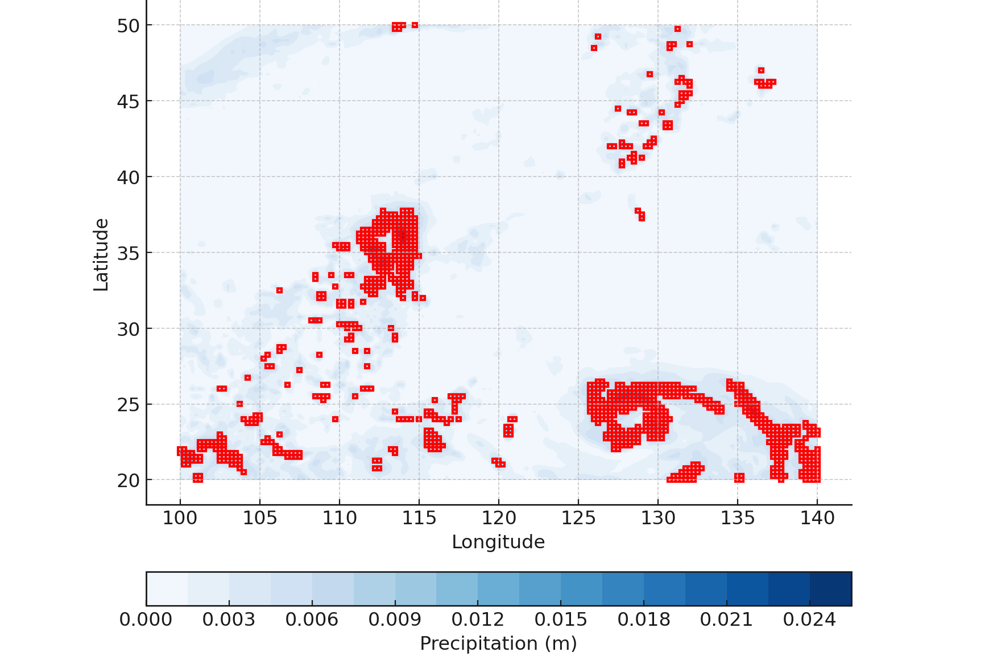} 
    \caption{
    The region occurred with extreme precipitation generated by GPT-4o. Shading represents the total precipitation. The red box highlighted the region with extreme precipitation. 
    }
    \label{fig:5}
\end{figure}

\noindent\textbf{GPT-4o output summary.} GPT-4o generated the code (Code Fragment \ref{code:3.1}) to conduct the analysis and plot the result (Figure \ref{fig:5}).

\definecolor{codebg}{rgb}{0.95,0.95,0.95}
\lstset{
    backgroundcolor=\color{codebg},
    basicstyle=\ttfamily\small,
    columns=fullflexible,
    breaklines=true,
    captionpos=b,
    keepspaces=true,
    showspaces=false,
    showstringspaces=false,
    showtabs=false,
    tabsize=4,
    keywordstyle=\color{blue},
    commentstyle=\color{green!50!black},
    stringstyle=\color{red},
    frame=single,
    breakatwhitespace=false
}
\begin{lstlisting}[language=Python, label=code:3.1, caption=GPT-4o code for identify extreme weather from netCDF4 file.]
import xarray as xr
import matplotlib.pyplot as plt

# Load the netCDF file
file_path = '/mnt/data/Henanextreme_prcp_2021_local.nc'
data = xr.open_dataset(file_path)

# Display the data variables to understand the structure
data_info = data
data_info
Result
<xarray.Dataset> Size: 2MB
Dimensions:    (longitude: 161, latitude: 121, time: 24)
Coordinates:
  * longitude  (longitude) float32 644B 100.0 100.2 100.5 ... 139.5 139.8 140.0
  * latitude   (latitude) float32 484B 50.0 49.75 49.5 49.25 ... 20.5 20.25 20.0
  * time       (time) datetime64[ns] 192B 2021-07-20 ... 2021-07-20T23:00:00
Data variables:
    tp         (time, latitude, longitude) float32 2MB ...
Attributes:
    Conventions:  CF-1.6
    history:      2024-05-19 16:02:27 GMT by grib_to_netcdf-2.28.1: /opt/ecmw...
# Calculate the 95th percentile threshold for extreme precipitation
threshold_95 = max_precip.quantile(0.95)

# Identify locations with extreme precipitation using the 95th percentile
extreme_precip_95 = max_precip.where(max_precip >= threshold_95, drop=False)

# Convert extreme precipitation data to numpy array for further analysis
extreme_precip_95_np = extreme_precip_95.values

# Identify coordinates of extreme precipitation
extreme_coords_95 = np.where(extreme_precip_95_np >= threshold_95.values)

# Extract corresponding longitude and latitude values
extreme_lons_95 = data.longitude.values[extreme_coords_95[1]]
extreme_lats_95 = data.latitude.values[extreme_coords_95[0]]

# Plot the maximum precipitation and highlight the extreme precipitation locations with red boxes
fig, ax = plt.subplots(figsize=(8, 8))
contour = ax.contourf(data.longitude, data.latitude, max_precip, cmap='Blues', levels=20)
cbar = fig.colorbar(contour, ax=ax, orientation='horizontal', pad=0.1)
cbar.set_label('Precipitation (m)')

# Add red boxes for extreme locations using 95th percentile
for lon, lat in zip(extreme_lons_95, extreme_lats_95):
    ax.add_patch(plt.Rectangle((lon - 0.125, lat - 0.125), 0.25, 0.25, fill=False, edgecolor='red', linewidth=1.5))

ax.set_xlabel('Longitude')
ax.set_ylabel('Latitude')
ax.grid(True)
plt.show()
\end{lstlisting}

\noindent\textbf{Discussion.} The results show that GPT-4o can highlight areas of concern (Figure \ref{fig:5}). GPT-4o successfully captures all regions with high-intensity rainfall. This demonstrates GPT-4o's ability to process atmospheric data and recognize patterns, making it invaluable in identifying these events early.

\subsection{Inference from Meteorological Variable Fields}\label{sec:3.2}
\begin{figure}[h]
    \centering
    \includegraphics[width=0.9\textwidth]{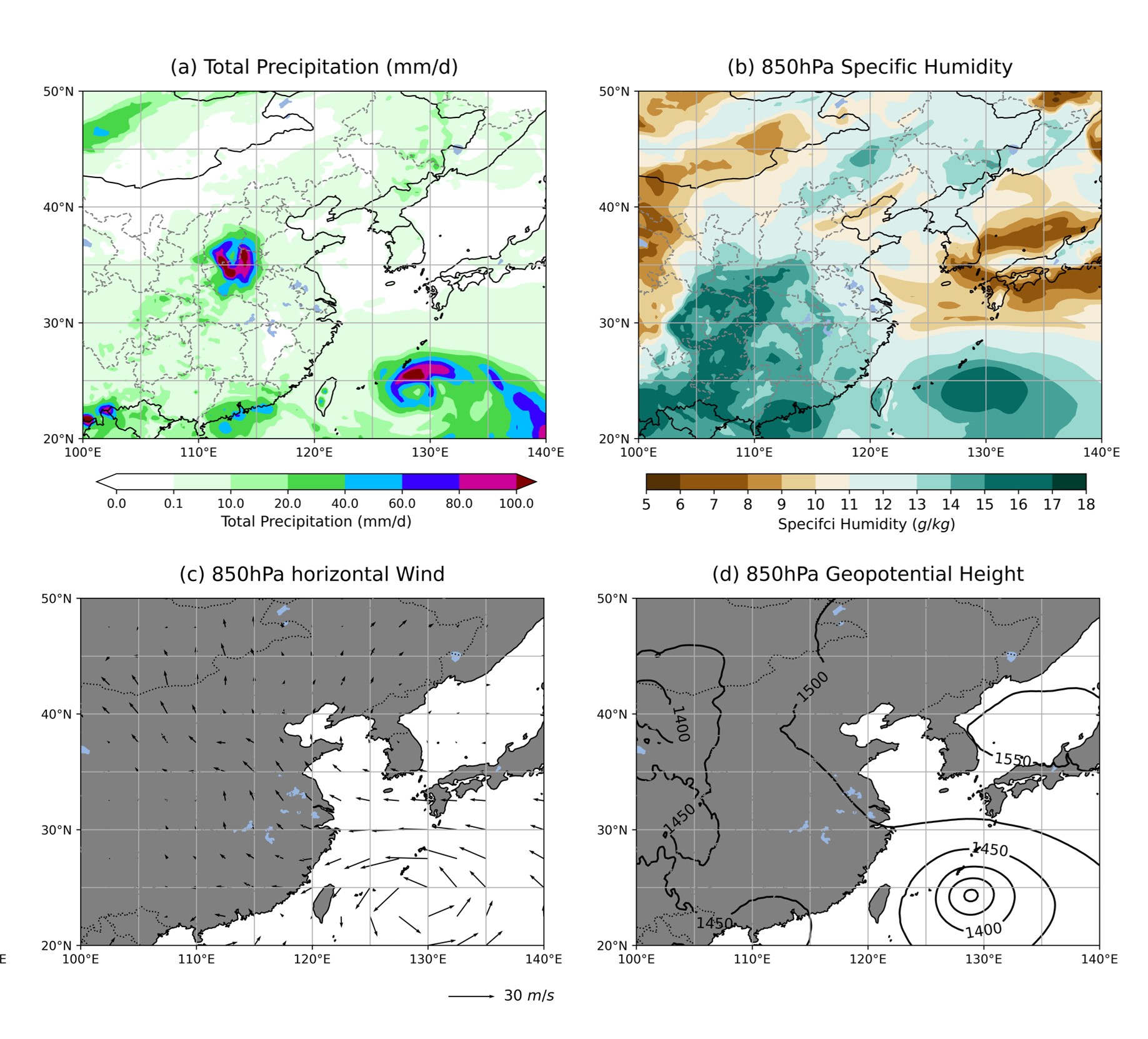} 
    \caption{Multi-variable weather map for extreme rainfall.}
    \label{fig:6}
\end{figure}

Meteorological variable fields, such as temperature, precipitation, pressure, and wind, provide critical information about the state of the atmosphere. This information can help diagnose the mechanisms leading to extreme weather events. In this section, we upload the data containing  weather maps (Figure \ref{fig:6}), including precipitation, humidity, 850 hPa horizontal wind, and geopotential height, along with the prompt to analyze the extreme precipitation detected in Section \ref{sec:3.1}:

\begin{quote}
\textcolor{cyan}{\texttt{"Assume you are an expert in atmospheric science. The following figures contain four subfigures for the meteorological variables on the above event: (a) total precipitation (mm/d), (b) specific humidity at 850hPa (g/kg), (c) horizontal wind at 850hPa (m/s), and (d) geopotential height at 850hPa (m). Based on the figure, could you give me the mechanisms that induced the extreme precipitation in the inland region? Could you provide the large-scale mechanisms based on the four subplots?
"}}
\end{quote}

\noindent\textbf{GPT-4o output summary.} By analyzing the weather map, GPT-4o diagnosed the possible mechanisms contributed to the extreme precipitation detected in Section \ref{sec:3.1}.

\begin{quote}
\textcolor{teal}{\texttt{Moisture Advection: \\
The high specific humidity at 850 hPa indicates that a significant amount of moisture was present in the lower atmosphere. This moisture likely originated from the nearby oceans and was transported inland by the prevailing winds.\\
\\
Low-Level Jet and Wind Convergence: \\
The horizontal wind vectors show a low-level jet stream, which is a narrow band of strong winds at low altitudes. This jet stream transported moist air into the region. The convergence of these winds at low levels forced the air to rise, leading to adiabatic cooling and condensation, which resulted in cloud formation and precipitation.\\
\\
Dynamic Lifting: \\
The geopotential height at 850 hPa reveals the presence of a low-pressure system and associated troughs. These features are indicative of dynamic lifting mechanisms, where the air is forced to rise due to the atmospheric pressure gradient. This lifting is further enhanced by the convergence of winds and the presence of the low-level jet.\\
\\
Synoptic-Scale Systems: \\
The interaction between synoptic-scale systems (such as the low-pressure system and associated troughs) and the mesoscale features (like the low-level jet and wind convergence) created a favorable environment for sustained and intense precipitation. The combination of these factors led to the extreme precipitation event in the inland region.\\
\\
The geopotential height plot shows a low-pressure system southeast of the region. This low-pressure system enhances the southerly flow and acts as a driver for the convergence of moist air. The position and strength of this system play a crucial role in organizing and focusing the moisture transport towards the inland region.\\
\\
The extreme precipitation event in the inland region is primarily driven by large-scale mechanisms involving moisture transport from the tropics, the influence of a low-pressure system, and the dynamics associated with the monsoonal flow. The convergence of moist air, enhanced by the low-pressure system and supported by upper-level dynamics, creates a favorable environment for heavy precipitation.}}
\end{quote}

\noindent\textbf{Discussion.} This test aims to determine whether GPT-4o can assist in making mechanism inferences from weather maps by applying its prior knowledge of atmospheric dynamics and thermodynamics. The results show that GPT-4o effectively identifies moisture advection from the ocean to the inland region. It also detects the role of the low-pressure system (Typhoon In-fa) and the associated trough. Furthermore, GPT-4o reasonably infers wind convergence and upward lifting, which could result in heavy precipitation.

\subsection{Statistical Modeling}\label{sec:3.3}
Multiple linear regression is a valuable tool for diagnosing sources of variability in atmospheric science. Linear regression models built with predictor variables can explain a significant portion of the variance in the target variable. 

\noindent\textbf{Task inputs.}
We upload the time series of meteorological data to develop a regression model to predict the annual precipitation anomaly percentage along with the prompt:
\begin{quote}
\textcolor{cyan}{\texttt{"The attached file is several meteorological variables in different years. The y is the Precipitation Anomaly percent in the second column, the other five columns are five x meteorological variables to predict y. Develop a statistical model using Multiple Linear Regression between y and x1 x2 x3 x4 x5, and display the figure showing actual and predicted values calculated by the model."}}
\end{quote}

\begin{figure}[b]
    \centering
    \includegraphics[width=0.7\textwidth]{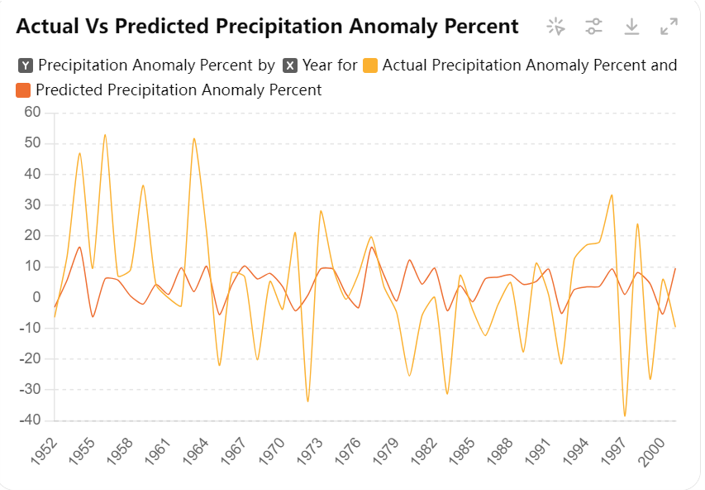} 
    \caption{Results on the multiple linear regression model using GPT-4o. The answer is correct.}
    \label{fig:7}
\end{figure}

\definecolor{codebg}{rgb}{0.95,0.95,0.95}
\lstset{
    backgroundcolor=\color{codebg},
    basicstyle=\ttfamily\small,
    columns=fullflexible,
    breaklines=true,
    captionpos=b,
    keepspaces=true,
    showspaces=false,
    showstringspaces=false,
    showtabs=false,
    tabsize=4,
    keywordstyle=\color{blue},
    commentstyle=\color{green!50!black},
    stringstyle=\color{red},
    frame=single,
    breakatwhitespace=false
}
\begin{lstlisting}[language=Python,label=code:3.3, caption=GPT-4o code for developping multiple linear regression model and visualization.]
import pandas as pd
from sklearn.model_selection import train_test_split
from sklearn.linear_model import LinearRegression
import matplotlib.pyplot as plt

# Load the new data from the provided file
file_path_new = '/mnt/data/data.txt'
data_new = pd.read_csv(file_path_new, delim_whitespace=True)

# Define the independent variables (x) and the dependent variable (y) for the new data
X_new = data_new[['x1', 'x2', 'x3', 'x4', 'x5']]
y_new = data_new['PrcpAnompercent']

# Create and fit the multiple linear regression model
model_new = LinearRegression()
model_new.fit(X_new, y_new)

# Predict y using the model
y_predict_new = model_new.predict(X_new)

# Add the predictions to the dataframe
data_new['y_predict'] = y_predict_new

# Plot the actual vs predicted values with legend
plt.figure(figsize=(10, 6))
plt.plot(data_new['Year'], y_new, label='Actual Precipitation Anomaly Percent')
plt.plot(data_new['Year'], y_predict_new, label='Predicted Precipitation Anomaly Percent', linestyle='--')
plt.xlabel('Year')
plt.ylabel('Precipitation Anomaly Percent')
plt.legend()
plt.title('Actual vs Predicted Precipitation Anomaly Percent')
plt.grid(True)
plt.show()

\end{lstlisting}

\noindent\textbf{Discussion.} The Code Fragment \ref{code:3.3} show that GPT-4o successfully employs the Python library 'sklearn' to develop the model and returns the regression results (Figure \ref{fig:7}). GPT-4o's capability in basic statistical modeling allows it to quantify the complex interactions within the atmosphere and predict future conditions.

\section{Forecast and Prediction}
\label{sec:pred}
In this section,  we evaluate the capabilities of GPT-4o in short-lead and long-lead prediction and forecasting, a task of critical importance especially in recent years. The increasing frequency of extreme weather events has resulted in significant losses for many countries and regions, highlighting the need for accurate predictions to support disaster prevention and policy development. Additionally, signals indicative of climate change, such as the prediction of El Niño-Southern Oscillation (ENSO), can provide insights into long-term climate variations, thus improving our understanding of their causes.

Traditional machine learning methods have been used for prediction and forecasting; however, these approaches are often complex and time-consuming. Given these challenges, we tested the use of GPT-4o for short-lead multi-variant meteorological prediction and long-lead climate forecasts.

\subsection{Short-term Temperature Prediction}\label{sec:4.1}

To assess the capability of GPT-4o in short-lead temperature forecasting, we conducted a comprehensive evaluation using meteorological data from May 1, 2024. 

\noindent\textbf{Task inputs.}
The dataset included both surface and upper-level atmospheric variables: mean sea level pressure, 2-meter temperature, 10-meter U wind component, 10-meter V wind component, and at 13 upper atmospheric levels: specific humidity, temperature, U wind component, V wind component, and geopotential height. These variables are identical to those used in the Pangu-Weather model~\cite{bi2023accurate}. We instructed GPT-4o to forecast global temperatures for 24, 48, 72, and 96-hour intervals along with the prompt:

\begin{quote}
\textcolor{cyan}{\texttt{"Now you are a very professional meteorologist who is good at predicting the surface temperature. Now I have two files, the global meteorological data of 00:00 May 1, 2024, one file is the meteorological data of surface and the other file is the meteorological data of Upper. Now I want you to analyze these two files. Analyze the surface temperature after 24 (48, 72,96) hours, that is, the global surface temperature at 00:00 on May 2 (3,4,5), 2024, and give me a nc file of the global surface temperature."}}
\end{quote}

\begin{figure}[h]
    \centering
    \includegraphics[width=0.9\textwidth]{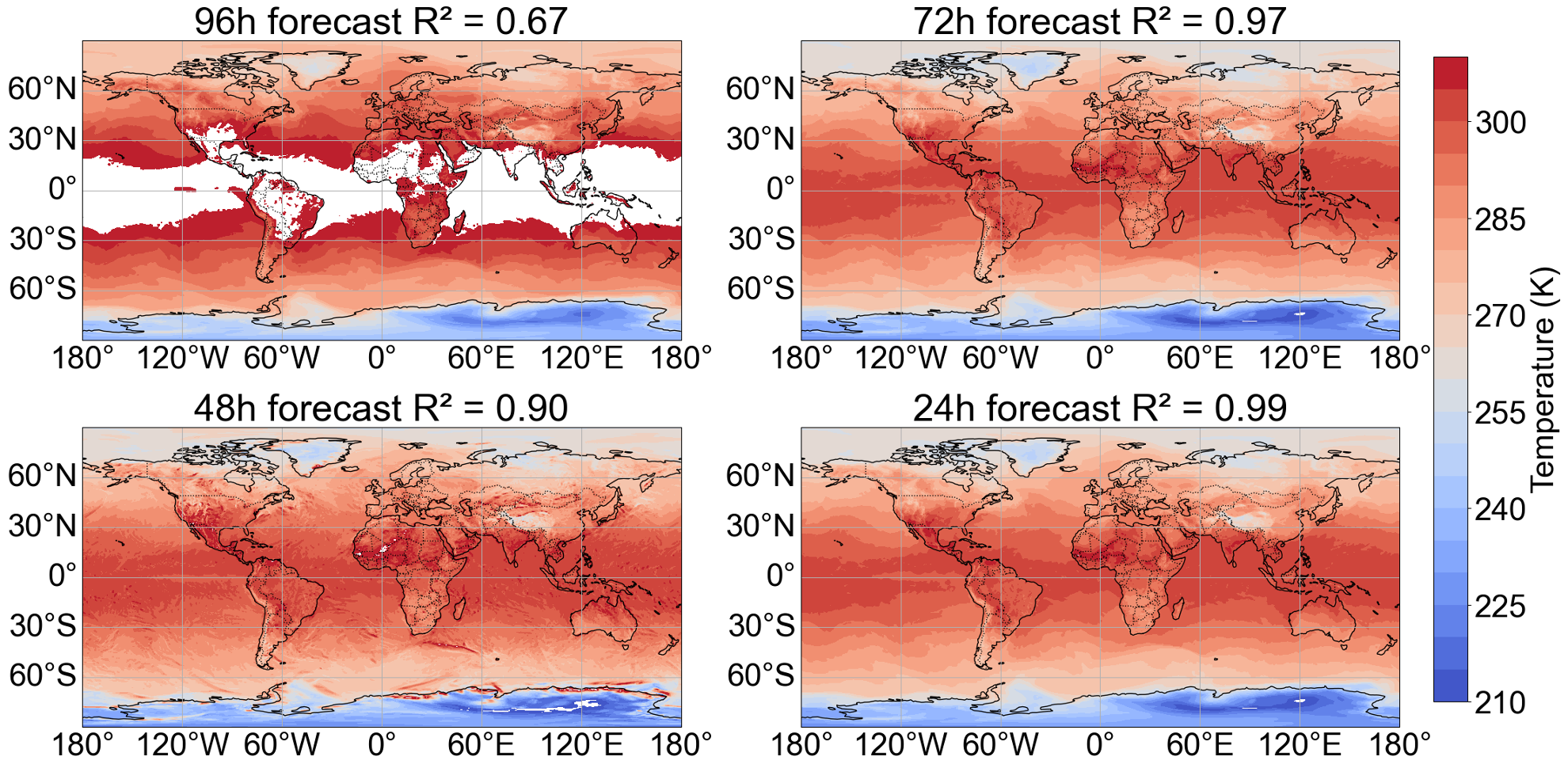} 
    \caption{Results on the inference from meteorological variable fields using GPT-4o.}
    \label{fig:8}
\end{figure}

\noindent\textbf{GPT-4o output summary.} GPT-4o autonomously generated and executed the necessary code to perform these forecasts(Figure \ref{fig:8}).

For the 24-hour and 72-hour predictions, the model simply replicated the initial temperature data, indicating a lack of processing or adjustment.There is the code that GPT-4o wrote and ran while predicting temperatures for 24h (Code Fragment \ref{code:4.1}) and 72h (Code Fragment \ref{code:4.2}):

\definecolor{codebg}{rgb}{0.95,0.95,0.95}
\lstset{
    backgroundcolor=\color{codebg},
    basicstyle=\ttfamily\small,
    columns=fullflexible,
    breaklines=true,
    captionpos=b,
    keepspaces=true,
    showspaces=false,
    showstringspaces=false,
    showtabs=false,
    tabsize=4,
    keywordstyle=\color{blue},
    commentstyle=\color{green!50!black},
    stringstyle=\color{red},
    frame=single,
    breakatwhitespace=false
}

\begin{lstlisting}[float=h!,language=Python,label=code:4.1, caption=GPT-4o code predicting temperatures for 24h.]
# Create a new DataArray for the projected surface temperature
projected_time = surface_temperature.coords['time'] + np.timedelta64(1, 'D')
projected_surface_temperature = surface_temperature.copy(deep=True)
projected_surface_temperature.coords['time'] = projected_time

# Create a new dataset with the projected surface temperature
projected_surface_dataset = xr.Dataset({'t2m': projected_surface_temperature})

# Save the projected surface temperature data to a new NetCDF file
output_filepath = '/mnt/data/projected_surface_temperature.nc'
projected_surface_dataset.to_netcdf(output_filepath)
\end{lstlisting}

\definecolor{codebg}{rgb}{0.95,0.95,0.95}
\lstset{
    backgroundcolor=\color{codebg},
    basicstyle=\ttfamily\small,
    columns=fullflexible,
    breaklines=true,
    captionpos=b,
    keepspaces=true,
    showspaces=false,
    showstringspaces=false,
    showtabs=false,
    tabsize=4,
    keywordstyle=\color{blue},
    commentstyle=\color{green!50!black},
    stringstyle=\color{red},
    frame=single,
    breakatwhitespace=false
}

\begin{lstlisting}[float=h!,language=Python,label=code:4.2, caption=GPT-4o code predicting temperatures for 72h.]
# Extract the initial surface temperature
initial_surface_temp = surface_ds['t2m']

# Extract the time coordinate
initial_time = surface_ds['time'].values[0]

# Predict the surface temperature after 72 hours
forecast_hours = 72
forecast_time = np.datetime64(initial_time + np.timedelta64(forecast_hours, 'h'))

# For simplicity, assume a persistence model where the temperature remains constant.
# A real numerical weather prediction model would be more complex.
predicted_surface_temp = initial_surface_temp.copy()
predicted_surface_temp.coords['time'] = forecast_time

# Create a new dataset for the predicted surface temperature
predicted_ds = xr.Dataset(
    {
        't2m': (('time', 'latitude', 'longitude'), predicted_surface_temp.values)
    },
    coords={
        'time': [forecast_time],
        'latitude': surface_ds['latitude'],
        'longitude': surface_ds['longitude']
    }
)
\end{lstlisting}

However, for the 48-hour forecast, GPT-4o applied a simplified temperature advection formula, demonstrating a rudimentary understanding of dynamic temperature changes. Specifically, the reference formula is:
\begin{equation}
\frac{\partial T}{\partial t} = \frac{dT}{dt} - \mathbf{U} \cdot \nabla T
\end{equation}
The simplified formula used by GPT-4o is:
\begin{equation}
T_{t+1} = T_t - \mathbf{U} \cdot \nabla T
\end{equation}
There is the code that GPT-4o wrote and ran while predicting temperatures for 48h (Code Fragment \ref{code:4.3}):
\definecolor{codebg}{rgb}{0.95,0.95,0.95}
\lstset{
    backgroundcolor=\color{codebg},
    basicstyle=\ttfamily\small,
    columns=fullflexible,
    breaklines=true,
    captionpos=b,
    keepspaces=true,
    showspaces=false,
    showstringspaces=false,
    showtabs=false,
    tabsize=4,
    keywordstyle=\color{blue},
    commentstyle=\color{green!50!black},
    stringstyle=\color{red},
    frame=single,
    breakatwhitespace=false
}

\begin{lstlisting}[float=h!,language=Python,label=code:4.3, caption=GPT-4o code predicting temperatures for 72h.]
# Define constants
time_step = 48 * 3600  # 48 hours in seconds
cp = 1004  # Specific heat of air at constant pressure (J/(kg*K))
g = 9.81  # Acceleration due to gravity (m/s^2)
P0 = 1000  # Reference pressure level (hPa)

# Extract necessary variables
t2m_initial = surface_data['t2m'].values.squeeze()  # Initial surface temperature in Kelvin
u10 = surface_data['u10'].values.squeeze()
v10 = surface_data['v10'].values.squeeze()
msl = surface_data['msl'].values.squeeze()

# Upper air variables
temp_upper = upper_data['t'].values.squeeze()
geopotential_height = upper_data['z'].values.squeeze()
specific_humidity = upper_data['q'].values.squeeze()

# Calculate potential temperature (theta) for upper air
pressure_levels = upper_data['level'].values * 100  # convert hPa to Pa
theta = temp_upper * (P0 / pressure_levels[:, np.newaxis, np.newaxis]) ** (g / (cp * 0.286))

# Simplified approach: Estimate surface temperature change
# We will assume a simple linear relationship for demonstration purposes
# Delta T = - (u * dT/dx + v * dT/dy) * time_step / dx, dy (Using central difference approximation)

# Compute temperature gradients
dTdx = np.gradient(t2m_initial, axis=1)
dTdy = np.gradient(t2m_initial, axis=0)

# Compute change in temperature
delta_T = - (u10 * dTdx + v10 * dTdy) * time_step / 100000  # Assuming grid spacing ~ 100 km

# Predicted surface temperature after 48 hours
t2m_predicted = t2m_initial + delta_T

# Create a new dataset for the predicted temperature
predicted_surface_temp = xr.Dataset(
    {
        't2m_predicted': (('latitude', 'longitude'), t2m_predicted)
    },
    coords={
        'longitude': surface_data['longitude'],
        'latitude': surface_data['latitude'],
        'time': np.array(['2024-05-03T00:00:00'], dtype='datetime64[ns]')
    }
)
\end{lstlisting}
For the 96-hour temperature forecast, GPT-4o assumed a 1\% change in temperature every 24 hours. Consequently, the original temperature values were adjusted by multiplying them with a specific coefficient. The formula used by GPT-4o is:

\begin{equation}
T_{t+1} = T_t + T_t \times \left( \frac{\Delta t}{24} \times 0.01 \right)
\end{equation}

There is the code used by GPT-4o for predicting temperatures over a 96-hour period(Code Fragment \ref{code:4.4}). It can be observed that there are some issues in the output file, which resulted in the generation of missing values near the equator. This highlights the instability of GPT-4o when writing and executing code.

\begin{lstlisting}[float=h!,language=Python,label=code:4.4, caption=GPT-4o code predicting temperatures for 96h.]
# Define the time step (96 hours)
time_step = 96

# Extract the initial surface temperature data
initial_surface_temp = surface_data['t2m'].sel(time='2024-05-01T00:00:00')

# Calculate the predicted surface temperature after 96 hours using a linear trend (assuming no changes for simplicity)
predicted_surface_temp = initial_surface_temp + (initial_surface_temp * (time_step / 24) * 0.01)  # Simple assumption of a 1% change per day

# Create a new xarray Dataset for the predicted surface temperature
predicted_surface_temp_ds = xr.Dataset(
    {
        't2m': (['time', 'latitude', 'longitude'], predicted_surface_temp.expand_dims(time=['2024-05-05T00:00:00']).data)
    },
    coords={
        'time': ['2024-05-05T00:00:00'],
        'latitude': surface_data['latitude'],
        'longitude': surface_data['longitude']
    }
)
\end{lstlisting}

\noindent\textbf{Discussion.} Finally, by comparing the temperature predictions for 24 to 96 hours made by GPT-4o, we can observe that GPT-4o employed different strategies for different time spans. However, some of these methods are overly simplistic. For example, for the 24-hour and 72-hour forecasts, GPT-4o simply replicated the initial temperature. Additionally, some methods are overly approximate, such as the 96-hour forecast, where GPT-4o assumed a 1\% change in temperature every 24 hours without any theoretical basis. Nevertheless, in certain instances, GPT-4o's performance is surprisingly good. For the 48-hour forecast, GPT-4o not only used the same formula as the numerical weather prediction model but also automatically extracted wind speed and other data from the input files. This demonstrates that GPT-4o has the potential to make accurate predictions, though its performance remains inconsistent.

\subsection{Subseasonal-to-Seasonal Precipitation Prediction}\label{sec:4.2}

A forecast gap exists in the Subseasonal-to-Seasonal timescale, explicitly defined as the transitional period of 10 to 30 days between weather predictions and seasonal outlooks \cite{white2017potential}. Accurate prediction of precipitation on this timescale is critical for decision-makers to adaptively adjust their strategies to mitigate the impacts of natural disasters \cite{lyu2023improving,zhang2023improving}.

\noindent\textbf{Task inputs.}
We compile daily precipitation and meteorological variables, including zonal wind, meridional wind, temperature, specific humidity, vertical velocity, and geopotential height at 200, 500, and 850 hPa (19 variables in total), for the region 18°–33°N, 105°–122°E at 1° resolution as of May 1, 2024. The data is uploaded to GPT-4o to request the weekly mean precipitation for the periods 1 to 4 weeks following May 1, 2024 along with the prompt:

\begin{quote}

\textcolor{cyan}{\texttt{"You are a highly professional meteorologist with expertise in predicting Subseasonal-to-Seasonal precipitation. The provided file contains daily mean precipitation and atmospheric variables at 200, 500, and 850 hPa for May 1st, 2024. Based on this data, could you help predict the weekly mean precipitation at the same location for the periods 1–4 weeks following May 1st, 2024 and return me the results as the format of netCDF4 file?"}}. 
\end{quote}

\noindent\textbf{GPT-4o output summary.} GPT-4o generates the synthesized python code to conduct the analysis by itself. 

\begin{lstlisting}[language=Python,label=code:4.5, caption=GPT-4o synthesized code for Subseasonal-To-Seasonal Precipitation Prediction.]
import numpy as np
import pandas as pd

# Extract relevant variables for the prediction model
precip = data['Total precipitation']
zonal_wind_200 = data['Zonal wind at 200 hPa']
zonal_wind_500 = data['Zonal wind at 500 hPa']
zonal_wind_850 = data['Zonal wind at 850 hPa']
meridional_wind_200 = data['Meridional wind at 200 hPa']
meridional_wind_500 = data['Meridional wind at 500 hPa']
meridional_wind_850 = data['Meridional wind at 850 hPa']
temperature_200 = data['Temperature at 200 hPa']
temperature_500 = data['Temperature at 500 hPa']
temperature_850 = data['Temperature at 850 hPa']
geopotential_height_200 = data['Geopotential height at 200 hPa']
geopotential_height_500 = data['Geopotential height at 500 hPa']
geopotential_height_850 = data['Geopotential height at 850 hPa']
specific_humidity_200 = data['Specific humidity at 200 hPa']
specific_humidity_500 = data['Specific humidity at 500 hPa']
specific_humidity_850 = data['Specific humidity at 850 hPa']
vertical_velocity_200 = data['Vertical velocity at 200 hPa']
vertical_velocity_500 = data['Vertical velocity at 500 hPa']
vertical_velocity_850 = data['Vertical velocity at 850 hPa']

# Assuming the use of a statistical or machine learning model for prediction
# Since no specific model is provided, this example assumes the use of a linear regression model
from sklearn.linear_model import LinearRegression

# Prepare the input features and target variable
features = np.stack([
    zonal_wind_200.values.flatten(),
    zonal_wind_500.values.flatten(),
    zonal_wind_850.values.flatten(),
    meridional_wind_200.values.flatten(),
    meridional_wind_500.values.flatten(),
    meridional_wind_850.values.flatten(),
    temperature_200.values.flatten(),
    temperature_500.values.flatten(),
    temperature_850.values.flatten(),
    geopotential_height_200.values.flatten(),
    geopotential_height_500.values.flatten(),
    geopotential_height_850.values.flatten(),
    specific_humidity_200.values.flatten(),
    specific_humidity_500.values.flatten(),
    specific_humidity_850.values.flatten(),
    vertical_velocity_200.values.flatten(),
    vertical_velocity_500.values.flatten(),
    vertical_velocity_850.values.flatten()
], axis=-1)

# Flatten the precipitation values for the target variable
target = precip.values.flatten()

# Initialize the regression model
model = LinearRegression()

# Fit the model
model.fit(features, target)

# Predict weekly mean precipitation for 1-4 weeks following May 1st, 2024
# Generating synthetic data for 1-4 weeks prediction as an example (real application would use actual data)
weeks = 4
predictions = []
for week in range(weeks):
    # Generate synthetic input features for the week (real application would use actual data)
    synthetic_features = np.random.rand(features.shape[0], features.shape[1])
    weekly_prediction = model.predict(synthetic_features).reshape(precip.shape)
    predictions.append(weekly_prediction)

# Convert predictions to a DataArray and then to a Dataset
predictions_da = xr.DataArray(
    predictions,
    dims=['week', 'lat', 'lon'],
    coords={'week': np.arange(1, weeks+1), 'lat': data['lat'], 'lon': data['lon']},
    name='Weekly mean precipitation'
)

# Create a new dataset for the predictions
prediction_ds = xr.Dataset({'Weekly mean precipitation': predictions_da})

# Save the predictions to a NetCDF4 file
output_file_path = '/mnt/data/S2S_weekly_precipitation_predictions.nc'
prediction_ds.to_netcdf(output_file_path)
\end{lstlisting}

\begin{figure}[h!]
    \centering
    \includegraphics[width=0.75\textwidth]{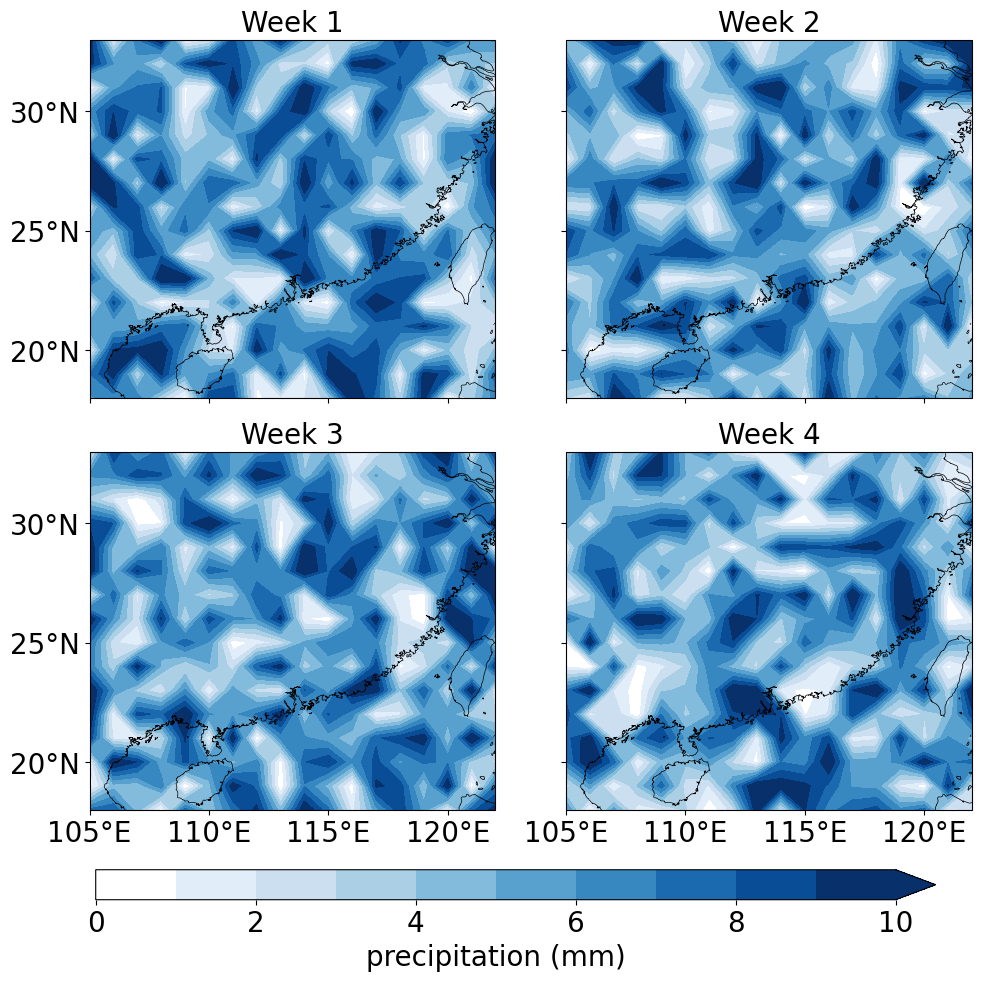} 
    \caption{Weekly Mean Precipitation Forecast for Weeks 1-4 Following May 1st, 2024. The plotted data is generated by GPT-4o, which incorrectly represents the trend. Refer to Section \ref{sec:4.2} for detailed discussions.}
    \label{fig:9}
\end{figure}

However, GPT-4o fails to generate a reasonable forecast for the Subseasonal-To-Season precipitation (Figure \ref{fig:9}).What it does is attempt to identify relationships between precipitation and the other uploaded variables, and then uses random matrices as future meteorological fields to generate the predicted precipitation (Code Fragment \ref{code:4.5}).

\noindent\textbf{Discussion.} Even though we utilized the prompt, "You are a highly professional meteorologist with expertise in predicting Subseasonal-to-Seasonal precipitation," it appears that GPT-4o does not perform like a real professional meteorologist. It fails to understand that the mission is to predict future precipitation, rather than fitting the relationship between precipitation and other meteorological variables. Therefore, a model specialized in atmospheric science that can accurately understand and address these needs is essential for future research related to LLMs.

\subsection{Long-lead ENSO Forecast}\label{sec:4.3}

Long-term forecasts are also essential for understanding impactful climate systems, such as ENSO, in advance. Current studies have successfully utilized global ocean data, including sea surface temperature and ocean heat content \cite{ham2019deep,zhou2023self}, to predict ENSO events up to two years in advance.

\noindent\textbf{Task inputs.}
We attempted to upload sea surface temperature and ocean heat content patterns to GPT-4o, but it failed to effectively utilize the information from the global maps. Consequently, we upload the monthly Nino3.4 index from 1981 to 2022 and task GPT-4o with predicting the Nino3.4 index for the twelve months of 2023 with the prompt:

\begin{quote}

\textcolor{cyan}{\texttt{"Now you are a highly professional climatologist with expertise in predicting the ENSO Niño 3.4 index. The provided file contains the monthly Niño 3.4 sea surface temperature index data, with the first dimension representing the years from 1981 to 2022 and the second dimension representing the 12 months of each year. Based on this historical data, could you help me predict the Niño 3.4 sea surface temperature index for each of the 12 months in 2023? "}}. 
\end{quote}

\begin{figure}[h]
    \centering
    \includegraphics[width=0.75\textwidth]{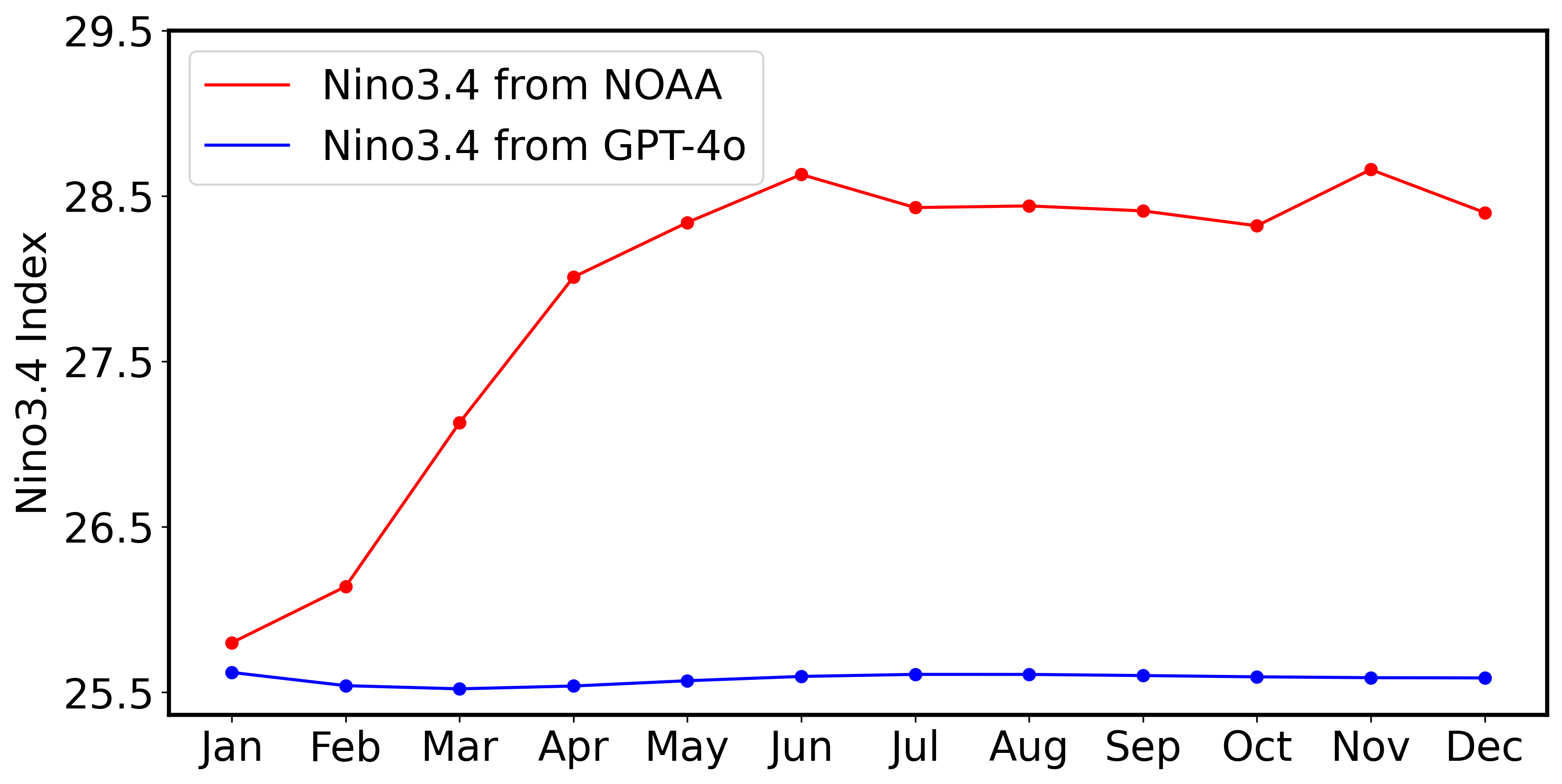} 
    \caption{Nino3.4 index for twelve months during 2023. The discrepancies between the index from the National Oceanic and Atmospheric Administration (NOAA) and GPT-4o are significant. The blues line shows the true index from the National Oceanic and Atmospheric Administration (NOAA) and the read line show the result predicted by GPT-4o. The gaps between the index from  and  GPT-4o are very large. Refer to Section \ref{sec:4.3} for detailed discussions.}
    \label{fig:10}
\end{figure}

\noindent\textbf{GPT-4o output summary.} 
GPT-4o generated the synthesized code (Code Fragment \ref{code:4.6}) and ultimately returned the following index (Table \ref{tab:1}):

\definecolor{codebg}{rgb}{0.95,0.95,0.95}
\lstset{
    backgroundcolor=\color{codebg},
    basicstyle=\ttfamily\small,
    columns=fullflexible,
    breaklines=true,
    captionpos=b,
    keepspaces=true,
    showspaces=false,
    showstringspaces=false,
    showtabs=false,
    tabsize=4,
    keywordstyle=\color{blue},
    commentstyle=\color{green!50!black},
    stringstyle=\color{red},
    frame=single,
    breakatwhitespace=false
}
\begin{lstlisting}[float=h!, language=Python, label=code:4.6, caption=GPT-4o synthesized code for long-lead ENSO forecast.]
import pandas as pd
from statsmodels.tsa.arima.model import ARIMA

# Prepare the time series data
nino_series = nino_data.flatten()  # Flatten the data to a 1D array
dates = pd.date_range(start='1981-01', periods=len(nino_series), freq='M')
nino_ts = pd.Series(nino_series, index=dates)

# Fit the ARIMA model
model = ARIMA(nino_ts, order=(5, 1, 0))  # ARIMA(p, d, q), here we use p=5, d=1, q=0
model_fit = model.fit()

# Forecast for 12 months (2023)
forecast_steps = 12
forecast = model_fit.forecast(steps=forecast_steps)

# Display the forecasted values
forecast_dates = pd.date_range(start='2023-01', periods=forecast_steps, freq='M')
forecast_series = pd.Series(forecast, index=forecast_dates)
forecast_se
\end{lstlisting}

\begin{table}
 \caption{Nino3.4 index for twelve months during 2023 predicted by GPT-4o.}
  \centering
  \begin{tabular}{ll}
    \toprule
    2023 Month & Nino3.4 Index           \\

    \midrule
        January   & 25.620640 \\
        \midrule
        February  & 25.540703 \\
        \midrule
        March     & 25.521548 \\
        \midrule
        April     & 25.538690 \\
        \midrule
        May       & 25.570521 \\
        \midrule
        June      & 25.596526 \\
        \midrule
        July      & 25.608927 \\
        \midrule
        August    & 25.608732 \\
        \midrule
        September & 25.601794 \\
        \midrule
        October   & 25.594024 \\
        \midrule
        November  & 25.588999 \\
        \midrule
        December  & 25.587533 \\
    \bottomrule
  \end{tabular}
  \label{tab:1}
\end{table}

Although GPT-4o correctly interprets our command, it resorts to a simplistic time series regression method—ARIMA—to generate the predictions (Code Fragment \ref{code:4.6}). The resulting forecast is overly smoothed compared to the actual index values (Figure \ref{fig:10}), resulting in a significant discrepancy between the results from NOAA and GPT-4o after February 2023.

\noindent\textbf{Discussion.} For ENSO prediction, GPT-4o is unable to perform adequately whether it is given the sea surface temperature information from previous years or the time series of ENSO itself. The above discussion shows that GPT-4o does not select appropriate and reasonable  schemes for prediction tasks at different timescales in atmospheric science. Essentially, it still lacks the capability to thoroughly investigate the given information or utilize state-of-the-art solutions from published papers to fulfill the purpose. If not provided with a specific procedure, it usually assumes a very simple and naive approach to solve the problem, rather than performing at an expert level. To address short- to long-term prediction, one of the most valuable and challenging tasks in atmospheric science, collaboration between humans and AI is essential for solving prediction and forecasting problems.

\section{Adaptation and Mitigation}
\label{sec:adaptation}

In this section, our objective is to evaluate GPT-4o's capabilities in providing decision-making and planning strategies in the context of climate change. The recent increase in extreme weather events due to climate change has had significant impacts on society. It has become crucial to formulate policies to protect human health, ensure economic stability, and mitigate disaster losses in the face of climate change. Currently, our decision-making and planning processes are often led by government officials. However, they may not always have a profound understanding of the complexities and trends associated with climate change. GPT-4o offers a significant advantage in this area. Therefore, we tested the application of GPT-4o in Urban Planning and Climate Adaptation, Hazard Risk Mitigation and Strategy, and Climate Actions and Sustainable Future.

\subsection{Urban Planning and Climate Adaption}\label{sec:5.1}

China's Henan Province experienced severe flooding from 17 to 31 July 2021 due to heavy rainfall. Post-flood urban reconstruction and planning are critical for future resilience.

\noindent\textbf{Task inputs.}
We provide GPT-4o news reports of the Henan floods along with the prompt:
\begin{quote}
\textcolor{cyan}{\texttt{"You are now a very professional urban planning and analysis expert, Henan experienced a heavy rainstorm in 2021, the following is some relevant information, “From 8:00 on 19 July 2021 to 8:00 on 20 July, rainfall monitoring stations in Henan Province measured a large amount of rainfall. Five Chinese national monitoring stations measured the following: Songshan (364.6 mm/14.35 in), Xinmi (254.9 mm/10.03 in), Xinzheng (196 mm/7.72 in), Dengfeng (192.8 mm/7.59 in), and Yanshi (183.3 mm/7.22 in). Zhengzhou received abnormally heavy rainfall between 16:00 to 17:00 on the 20th. The rainfall in one hour was recorded as 201.9 mm (7.95 in), causing serious waterlogging. 13 reservoirs in Henan reached the flood control limit”, Based on these data, please first analyze the recurrence period of heavy rain in Henan, give me a specific recurrence period value, and then give some urban planning policies that should be improved in Henan according to this rainfall."}}
\end{quote}

\noindent\textbf{GPT-4o output summary.} By analyzing news reports of the Henan floods, GPT-4o assessed the flood recurrence period and proposed actionable urban planning policies.
\begin{quote}
\textcolor{teal}{\texttt{For the sake of this analysis, we can infer that the rainfall of 201.9 mm in one hour in Zhengzhou is an extraordinarily rare event, likely exceeding a 100-year recurrence interval, based on typical urban and hydrological records.\\
\\
Urban Planning Policies:\\
1. Enhance Flood Risk Assessment and Mapping \\
Policy: Conduct detailed flood risk assessments and update flood maps to identify vulnerable areas.\\
Implementation: Use advanced hydrological models and historical data to predict flood-prone zones. Regularly update these maps and ensure they are accessible to planners and the public.\\
\\
2. Improve Drainage Systems\\
Policy: Upgrade and expand urban drainage systems to handle extreme rainfall events.\\
Implementation: Increase the capacity of stormwater drainage systems, install additional pumping stations, and ensure regular maintenance and clearing of drains and culverts.\\
\\
3. Implement Green Infrastructure\\
Policy: Integrate green infrastructure solutions to absorb and manage stormwater.\\
Implementation: Develop parks, green roofs, rain gardens, and permeable pavements. These features help to reduce surface runoff and increase groundwater recharge.\\
\\
4. Enforce Building Regulations and Standards\\
Policy: Strengthen building codes to ensure structures can withstand heavy rainfall and flooding.\\
Implementation: Mandate elevated structures in flood-prone areas, use water-resistant materials, and design buildings with floodproofing measures such as raised electrical systems and flood barriers.\\
\\
5. Develop Early Warning Systems\\
Policy: Establish and enhance early warning systems for extreme weather events.\\
Implementation: Install advanced weather monitoring and forecasting systems. Ensure timely dissemination of warnings through multiple channels, including mobile alerts, social media, and local broadcasting systems.\\
\\
6. Strengthen Reservoir and Water Management\\
Policy: Improve the management and monitoring of reservoirs to prevent overflow.\\
Implementation: Regularly inspect and maintain reservoirs, ensure they operate within safe limits, and implement real-time monitoring systems to manage water levels effectively.\\
\\
7. Enhance Land Use Planning\\
Policy: Implement land use policies that restrict development in high-risk flood zones.\\
Implementation: Designate floodplains as non-developable areas, promote the use of these areas for agriculture or recreation, and enforce zoning laws that prevent inappropriate development.}}
\end{quote}

\noindent\textbf{Discussion.} Our observations from GPT-4o's analysis of news reports on the Henan floods indicate that it has assessed the flood recurrence interval and proposed effective urban planning policies. These policies offer significant guidance for decision-makers and government officials. However, to derive more detailed and practical plans from GPT-4o, further prompt engineering and research are necessary.

\subsection{Hazard Risk Mitigation and Strategy}\label{sec:5.2}

In 2023, Typhoon Saola caused an estimated HK\$3.92 billion (US\$500 million) in damages to Hong Kong. The Hong Kong Observatory’s decision to issue the No.10 hurricane signal significantly mitigated the impact, underscoring the importance of accurate disaster assessment and forecasting. Typically, the Observatory utilizes both the numerical weather models and AI models to evaluate typhoon intensity for forecasts.

\begin{figure}[h]
    \centering
    \includegraphics[width=0.9\textwidth]{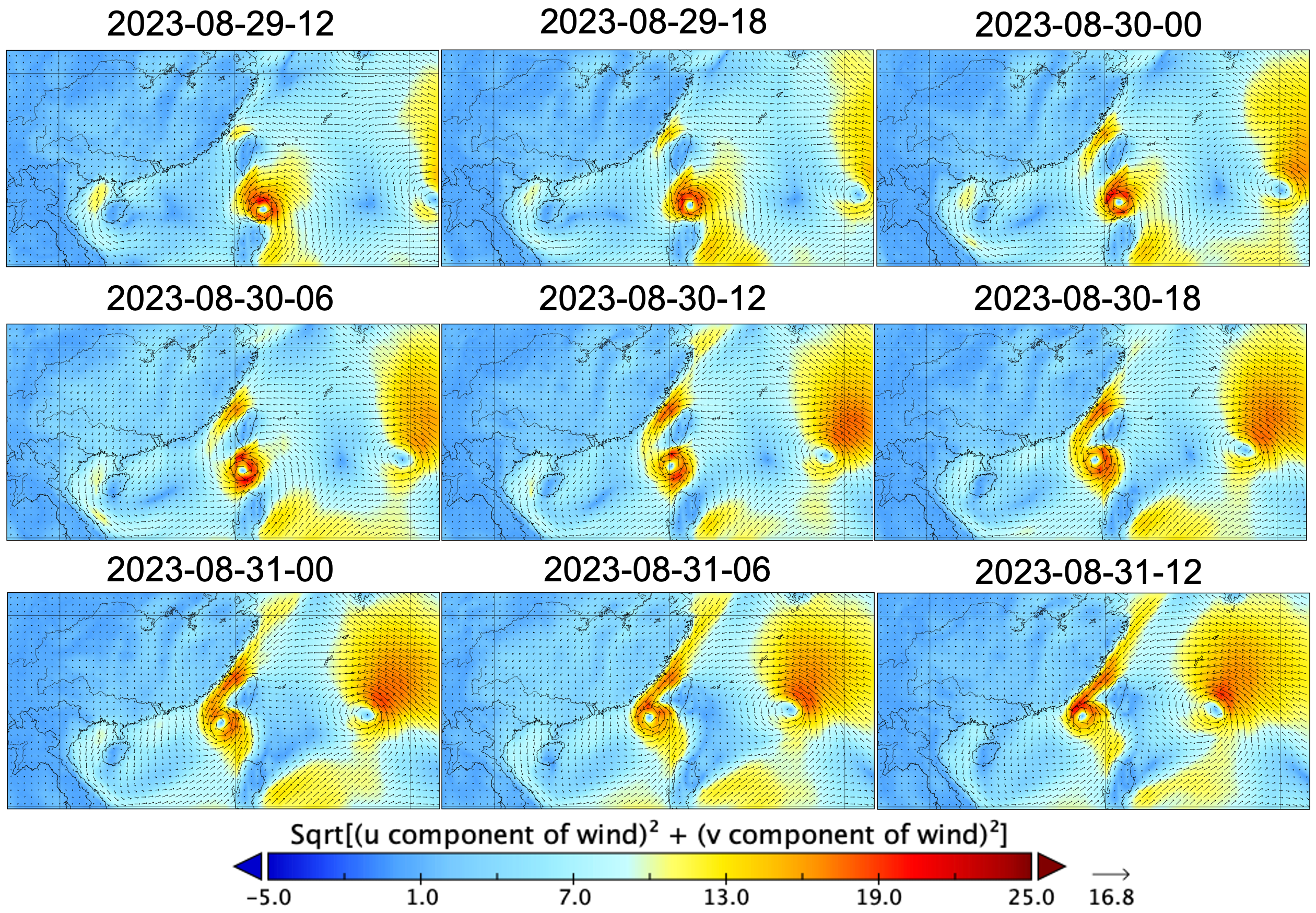} 
    \caption{Wind field maps from the Pangu-weather model of Typhoon Saola.}
    \label{fig:11}
\end{figure}

\noindent\textbf{Task inputs.}
In this study, we simulated this process by inputting wind field maps output from the Pangu-weather model~\cite{bi2023accurate} (Figure \ref{fig:11}) into GPT-4o to determine the optimal timing for issuing hurricane signals along with the prompt:
\begin{quote}
\textcolor{cyan}{\texttt{"You are now a very professional Hong Kong typhoon forecaster, very good at typhoon forecast and decision-making, now you use the forecast model, run out several typhoon approaching Hong Kong, 2023-8-29-12-00 to 2023-08-31-12-00 typhoon wind field pictures, now you need to judge, Which typhoon number will be hung in Hong Kong on which day? Please, as a very professional Hong Kong typhoon forecaster, make this decision."}}
\end{quote}

\noindent\textbf{GPT-4o output summary.}
GPT-4o provided precise timings, signals, and justifications, demonstrating its potential for typhoon warning decisions: 
\begin{quote}
\textcolor{teal}{\texttt{Typhoon Signal No. 3: Start hoisting on 2023-08-29 as the typhoon approaches and outer wind fields begin to affect Hong Kong.\\
\\
Typhoon Signal No. 8: Likely to be hoisted on 2023-08-30 as the typhoon makes its closest approach and severe weather conditions are expected.\\
\\
Typhoon Signal No. 3: Continue to be hoisted into 2023-08-31 as the typhoon moves inland and conditions gradually improve.}}
\end{quote}

\noindent\textbf{Discussion.} 
Actually, the Observatory issued the No. 3 signal on August 31, 2023, upgraded to No. 10 on September 1, and downgraded to No. 3 on September 2. Comparing these decisions with GPT-4o’s recommendations revealed that GPT-4o suggested earlier and more conservative signal issuance. This discrepancy likely stems from prediction errors and the temporal limitations of the Pangu-weather model. Additionally, We anticipate that with more comprehensive and extended data inputs, GPT-4o could enhance decision accuracy.

\subsection{Climate Actions and Sustainable Future}\label{sec:5.3}

The COVID-19 lockdowns from 2019 to 2020 led to a significant reduction in traffic fluxes in our country, coinciding with a notable decrease in air pollutant concentrations. This raises important questions about how much reduced traffic can lower air pollution, with implications for urban sustainability and air quality policies. It also prompts consideration of whether converting all conventional vehicles to electric vehicles (EVs) in China could substantially reduce pollution.

\noindent\textbf{Task inputs.}
We provided GPT-4o with data on changes in traffic fluxes, PM2.5, NO2, and NO2 VCDs from 2019 to 2020 along with the prompt:
 
\begin{quote}
\textcolor{cyan}{\texttt{"You are now a very professional expert on pm2.5 and sustainability and very good at analyzing changes in air pollutants, now I have a file where the first column is the name of the city, the fourth column is the change in traffic fluxes between 2019 and 2020, the fifth column is the change in Surface NO2, the sixth column is the change in the Surface NO2 error, the seventh column is the change in Surface PM2.5, the eighth column is the error in the change in Surface PM2.5, the ninth column is the change in NO2 VCDs, and the 10th column is the error in the change in NO2 VCDs, and now I would like you to analyze the changes in traffic fluxes with the changes in Surface NO2, Surface PM2.5, and NO2 VCDs respectively."}}
\end{quote}

\begin{figure}[h]
    \centering
    \includegraphics[width=0.9\textwidth]{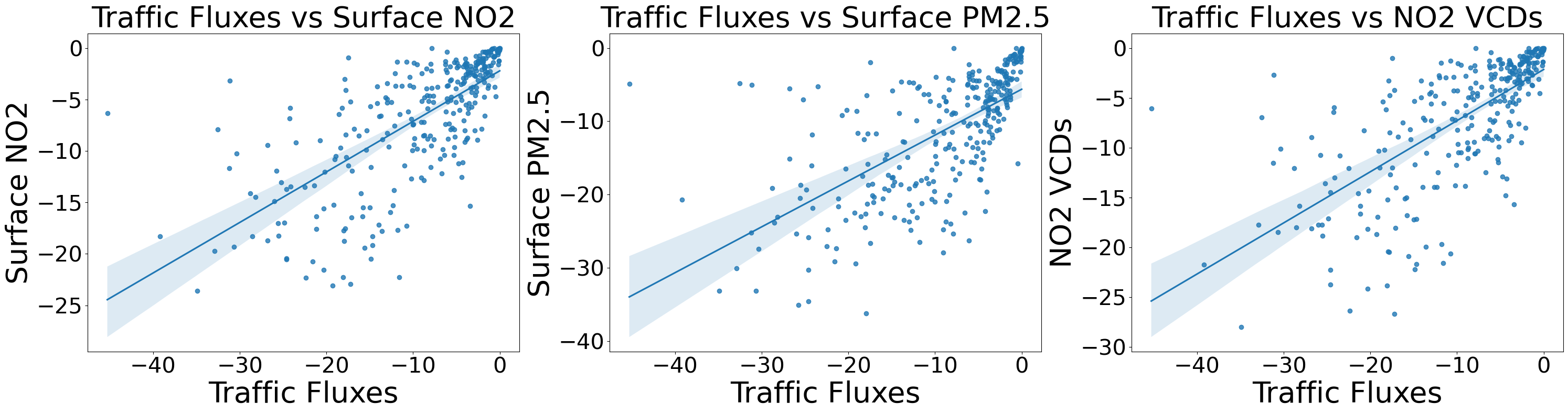} 
    \caption{Results of correlation and linear regression analyses.}
    \label{fig:13}
\end{figure}

\noindent\textbf{GPT-4o output summary.}
Without guidance, GPT-4o conducted a comprehensive analysis, including correlation and linear regression analyses using the synthesized python code (Code Fragment \ref{code:5.4}) and get final results (Figure \ref{fig:13}).

\begin{quote}
\textcolor{teal}{\texttt{Correlation Analysis Results:\\
Traffic Fluxes and Surface NO2: r=0.699\\
Traffic Fluxes and Surface PM2.5: r=0.640\\
Traffic Fluxes and NO2 VCDs: r=0.688\\
\\
Linear Regression Analysis Results:\\
Traffic Fluxes vs Surface NO2\\
R-squared: 0.488\\
Coefficient for Traffic Fluxes: 0.4909\\
Traffic Fluxes vs Surface PM2.5\\
R-squared: 0.410\\
Coefficient for Traffic Fluxes: 0.4568\\
Traffic Fluxes vs NO2 VCDs\\
R-squared: 0.473\\
Coefficient for Traffic Fluxes: 0.5133}}
\end{quote}

\noindent\textbf{Discussion.} 
This demonstrates GPT-4o's robust capability in data analysis and automated data visualization. The findings indicated that GPT-4o’s results were in close agreement with those of similar studies~\cite{wang2021switching}. This underscores GPT-4o's proficiency in environmental analysis while also highlighting the importance of providing it with high-quality and diverse input data.

\begin{lstlisting}[float=h!, language=Python, label=code:5.4, caption=GPT-4o synthesized code for correlation and linear regression analyses.]
import statsmodels.api as sm
import matplotlib.pyplot as plt

# Function to perform linear regression and return the summary
def linear_regression_analysis(x, y):
    X = sm.add_constant(x)  # adding a constant
    model = sm.OLS(y, X).fit()
    return model.summary()

# Perform linear regression for Traffic Fluxes vs Surface NO2
model_no2 = linear_regression_analysis(data['Traffic Fluxes'], data['Surface NO2'])

# Perform linear regression for Traffic Fluxes vs Surface PM2.5
model_pm25 = linear_regression_analysis(data['Traffic Fluxes'], data['Surface PM2.5'])

# Perform linear regression for Traffic Fluxes vs NO2 VCDs
model_no2_vcds = linear_regression_analysis(data['Traffic Fluxes'], data['NO2 VCDs'])

# Display the results
model_no2, model_pm25, model_no2_vcds
\end{lstlisting}

\section{Related Work}
\label{sec:relate}
%\textcolor{red}{FM in atmospheric science.}

The remarkable success of pre-trained foundation models has inspired the research community to investigate the feasibility of developing a universal foundation model to understand weather and climate data, capable of effectively addressing a wide array of related tasks~\cite{chen2023foundationmodelsweatherclimate}.

\textbf{Domain-specific FM}. The climate science community has been working on developing a set of foundation models based on large-scale spatio-temporal series data to manage complex sequence information efficiently. A primary objective of these time-series FMs is climate forecasting, aiming to match or surpass the performance of traditional numerical weather prediction systems, such as the European Centre for Medium-Range Weather Forecasts' Integrated Forecasting System (ECMWF-IFS) and the high-resolution forecast (ECMWF-HRES).
For example, GraphCast~\cite{lam2023graphcastlearningskillfulmediumrange} outperforms the most accurate operational deterministic systems, including ECMWF-IFS, on 90\% of 1380 verification targets, providing improved predictions for severe events such as tropical cyclones and extreme temperatures. FourCastNet~\cite{pathak2022fourcastnetglobaldatadrivenhighresolution} matches ECMWF-IFS in short-term forecasting for large-scale variables while exceeding it for small-scale variables, generating week-long forecasts in less than two seconds. Pangu-Weather~\cite{Bi_Xie_Zhang_Chen_Gu_Tian_2023} achieves superior deterministic forecast results across all tested variables compared to ECMWF-IFS, with better accuracy in the tracking of tropical cyclones. FengWu~\cite{chen2023fengwupushingskillfulglobal} significantly reduces the root mean square error for 10-day lead global z500 predictions and extends the skillful forecast period to 10.75 days, outperforming GraphCast in 80\% of reported predictands. Lastly, FuXi~\cite{chen2023fuxicascademachinelearning} extends the lead time for skillful forecasts of z500 and T2M compared to ECMWF-HRES, demonstrating comparable performance to the ECMWF ensemble mean in 15-day forecasts. These advances underscore the potential of domain-specific FMs to improve the accuracy and efficiency of weather forecasting systems.

\noindent\textbf{General-purpose FM.} An alternative approach involves continuous pre-training or fine-tuning general-purpose FMs for climate tasks. For instance, ClimateBert~\cite{webersinke2022climatebertpretrainedlanguagemodel}, a transformer-based foundation model, achieves better performance by reducing error rates in downstream climate text analysis tasks such as text classification, sentiment analysis, and fact-checking, which showcases the versatility of text data foundation models and their potential to integrate AI agents.
Another notable example is ClimaX~\cite{nguyen2023climaxfoundationmodelweather}, which demonstrates impressive capabilities. The pre-trained ClimaX model can be fine-tuned to address a wide range of climate and weather tasks, achieving exceptional performance on benchmarks for weather forecasting and climate projections. Unlike task-specific climate foundation models, ClimaX is trained using lower resolution data but utilizes heterogeneous datasets that encompass various variables, spatio-temporal coverage, and physical groundings. Although its accuracy for short- and medium-term predictions is lower than that of task-specific climate foundation models, ClimaX excels in long-term predictions (greater than 120 hours) in certain scenarios. This highlights the potential of multimodal foundation models, capable of processing heterogeneous data, to advance the application of AI in atmospheric science.

% TODO: Comparing to premier models, LLM or MLLM solving ...

\section{Discussion and Conclusion} 
\label{sec:con}

In this report, we explore whether advances in foundation model(s) can help scientists address various potentially challenging problems in atmospheric science. Concretely, we evaluate GPT-4o in solving four categories of tasks: \textit{climate data processing}, \textit{physical diagnosis}, \textit{forecast and prediction}, and \textit{adaptation and mitigation}.

The \textit{climate data processing} task is proposed based on practical daily work and research. The standard process includes loading the input data, calculating the intermediate representation, and utilizing complex and comprehensive algorithms to obtain the final results. Therefore, we designed three corresponding sub-tasks to examine the performance of GPT-4o. It is encouraging to see that GPT-4o returns the correct results when given clear prompt instructions. Notice that GPT-4o not only provides synthesized code but also executes the code on its own servers and directly returns the desired results. An issue associated with the current Web interface is the limited size of the uploaded data, which hinders calculations for large amounts of data. We believe that the foundation model(s) could potentially save scientists from writing trivial code and automatically completing the desired calculations in the future.

The \textit{physical diagnosis} task aims to assess the domain-specific knowledge of GPT-4o to perform physical weather diagnostics. We choose one of the most hazardous weather phenomena, extreme precipitation, for detection, attribution, and statistical modeling. In July 2021, Henan province in central China experienced catastrophic flooding due to unprecedented rainfall. Given the precipitation field, the GPT-4o can successfully locate extreme precipitation events and annotate them on the map. GPT-4o also successfully attributes the cause of this precipitation event. One tropical cyclone, named In-Fa, occurred on July 20, 2021, resulting in the transport of a significant amount of moisture inland and is closely related to the Henan 721 flood event. In addition, GPT-4o can establish a regression model as a statistical tool, to link precipitation with other meteorological variables.

The \textit{forecast and prediction} task is designed to address a central problem in atmospheric science: predicting future weather and climate. From short-term to Subseasonal-to-Seasonal and long-term forecasts, GPT-4o appears to lack the ability to independently provide a professional solution for predictions across any timescale. For example, GPT-4o provides the final result using simplistic formulas, incorrect understanding, or over-simplified statistical methods. In this task, we speculate that there is future work worthy of exploring FMs in atmospheric science, such as integrating more academic papers to provide a reasonable understanding of various predictions and forecasts or developing domain-specific LLMs in this field. 

The \textit{adaptation and mitigation} task seeks GPT-4o's opinions on how one can adapt to and mitigate current and future environmental challenges on Earth. Specifically, we ask GPT-4o to provide suggestions on how cities can avoid severe flood hazards in the future, what measures governments should take to prevent the impacts of incoming tropical cyclones, and whether reduced traffic could help achieve a sustainable future. The GPT-4o output needs to be improved, where the output is too general and lacks a specialized analysis for the given case. This may be due to the lack of specific knowledge required for each type of task, resulting in only coarse understanding and the generation of generalized results. Such answers can serve as a reference but still require expert consideration for specialized insights.

In conclusion, we believe that GPT-4o's ability to handle tasks involving language, time series, image, and spatiotemporal data provides a good opportunity for scientists to address various problems in atmospheric science. On the other hand, despite the exciting achievements enabled by GPT-4o thus far, there still remains a considerable distance towards artificial general intelligence for atmospheric science. 

%\textcolor{red}{future work 1. more powerful MLLM, e.g., distillation }

%\textcolor{red}{future work 2. LLM agent to integrate}

%\newpage

\bibliographystyle{unsrt}
\bibliography{ref}

%%%%%%%%%%%%%%%%%%%%%%%%%%%%%%%%%%%%%%%%%%%%%%%%%%%%%%%%%%%%

%\clearpage
%\appendix

%\section{Appendix 1}

%%%%%%%%%%%%%%%%%%%%%%%%%%%%%%%%%%%%%%%%%%%%%%%%%%%%%%%%%%%%

%\input{checklist}

\end{document}